\pdfoutput=1
\documentclass[10pt,journal,compsoc]{IEEEtran}
\ifCLASSOPTIONcompsoc
  \usepackage[nocompress]{cite}
\else
  \usepackage{cite}
\fi
\ifCLASSINFOpdf
  \usepackage[pdftex]{graphicx}
\else
  \usepackage[dvips]{graphicx}
\fi

\usepackage{epsfig}
\usepackage{graphicx}
\usepackage{amsmath}
\usepackage{array}
\usepackage{amssymb}
\usepackage{enumitem}
\usepackage{tikz}

\usepackage[pagebackref=true,breaklinks=true,letterpaper=true,colorlinks,bookmarks=false,urlcolor=red,citecolor=cyan]{hyperref}
\usepackage{gensymb,graphics,subfigure,inputenc}
\usepackage[linesnumbered,algo2e,boxed]{algorithm2e}
\graphicspath{{Figure/}}
\usepackage[figuresright]{rotating}
\usepackage{amsmath,amssymb,textcomp,subfigure,multirow,upgreek}
\usepackage{booktabs}
\usepackage{color,mdwlist}

\usepackage{ragged2e}

\usepackage{review}

\usepackage{graphicx,fontawesome}
\graphicspath{{Figures/}}

\usepackage{caption}
\usepackage{wrapfig}

\usepackage{enumitem}
\setitemize{noitemsep,topsep=0pt,parsep=0pt,partopsep=0pt}

\begin{document}
%
\title{Multimodal Perception for Goal-oriented Navigation: A Survey}
\author{I-Tak Ieong\quad Hao Tang$^*$
	\IEEEcompsocitemizethanks{
	    \IEEEcompsocthanksitem I-Tak Ieong is with the School of Computer Science and Technology, Tongji University, Shanghai 201804, China. E-mail: 2250268@tongji.edu.cn \protect
	    \IEEEcompsocthanksitem Hao Tang is with the School of Computer Science, Peking University, Beijing 100871, China. E-mail: haotang@pku.edu.cn \protect
        }
	\thanks{$^*$Corresponding author: Hao Tang.}
}

%
%

\markboth{Submitted IEEE Transactions on Pattern Analysis and Machine Intelligence}%
{Shell \MakeLowercase{\textit{et al.}}: Bare Demo of IEEEtran.cls for Computer Society Journals}
%



\IEEEtitleabstractindextext{%
\justify
\begin{abstract}

    Goal-oriented navigation presents a fundamental challenge for autonomous systems, requiring agents to navigate complex environments to reach designated targets. This survey offers a comprehensive analysis of multimodal navigation approaches through the unifying perspective of inference domains, exploring how agents perceive, reason about, and navigate environments using visual, linguistic, and acoustic information. Our key contributions include organizing navigation methods based on their primary environmental reasoning mechanisms across inference domains; systematically analyzing how shared computational foundations support seemingly disparate approaches across different navigation tasks; identifying recurring patterns and distinctive strengths across various navigation paradigms; and examining the integration challenges and opportunities of multimodal perception to enhance navigation capabilities. In addition, we review approximately 200 relevant articles to provide an in-depth understanding of the current landscape.
\end{abstract}

}

\maketitle

\IEEEdisplaynontitleabstractindextext

%
\IEEEpeerreviewmaketitle


%
%
%
%

\section{Introduction}
\IEEEPARstart{E}mbodied navigation represents a fundamental challenge in autonomous systems—enabling agents to traverse unknown environments to reach specified targets. During the past decade, significant advances have transformed navigation from simple geometric path planning \cite{yamauchi1997frontier,sethian1996fast} to sophisticated multimodal reasoning that integrates visual \cite{he2016deep,vaswani2017attention}, linguistic \cite{brown2020language,achiam2023gpt}, and audio information \cite{chen2020soundspaces,morgado2020learning}. As the field matures, successful approaches increasingly bridge low-level perception with high-level semantic understanding through diverse computational frameworks, which we term inference domains.

Moreover, while varying in goal specification methods (coordinates in PointNav, visual targets in ImageNav, semantic categories in ObjectNav \cite{anderson2018evaluation,batra2020objectnav}, and audio sources in AudioGoalNav \cite{chen2020soundspaces}), share common underlying mechanisms for environmental perception, representation, and decision-making. The field has evolved from early modular approaches with explicit mapping and planning components \cite{chaplot2020learning,yamauchi1997frontier} to modern end-to-end architectures powered by deep learning \cite{yokoyama2024hm3d,ramrakhya2023pirlnav}. This evolution reflects a transition from purely geometric reasoning to sophisticated semantic understanding, and there is a shift visible across all navigation paradigms.

\begin{figure*}[t]
    \centering
    \begin{tikzpicture}
        \node[anchor=south west,inner sep=0] (image) at (0,0) {
            \includegraphics[width=\textwidth]{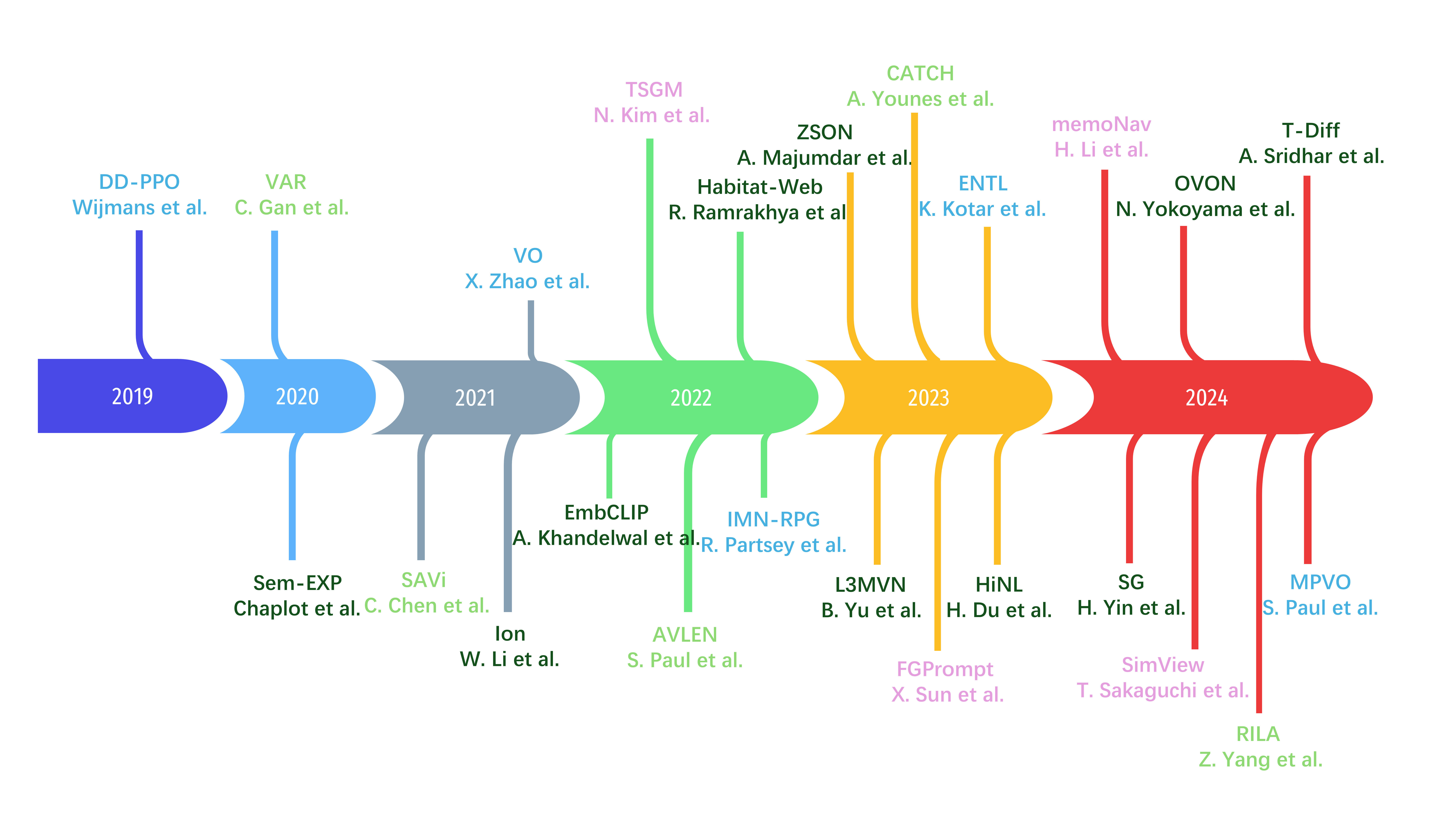}
        };
        
        \begin{scope}[x={(image.south east)},y={(image.north west)}]
            \node[inner sep=5pt,rectangle,fill=none,draw=none] at (0.10,0.81) {
                \hbox{\cite{wijmans2019dd}\phantom{.}}
            }; 
            \node[inner sep=5pt,rectangle,fill=none,draw=none] at (0.36,0.72) {
                \hbox{\cite{zhao2021surprising}\phantom{.}}
            };
            \node[inner sep=5pt,rectangle,fill=none,draw=none] at (0.55,0.40) {
                \hbox{\cite{partsey2022mapping}\phantom{.}}
            };
            \node[inner sep=5pt,rectangle,fill=none,draw=none] at (0.68,0.81) {
                \hbox{\cite{kotar2023entl}\phantom{.}}
            };
            \node[inner sep=5pt,rectangle,fill=none,draw=none] at (0.91,0.23) {
                \hbox{\cite{paul2024mpvo}\phantom{.}}
            };
            
            \node[inner sep=5pt,rectangle,fill=none,draw=none] at (0.20,0.22) {
                \hbox{\cite{chaplot2020object}\phantom{.}}
            };
            \node[inner sep=5pt,rectangle,fill=none,draw=none] at (0.35,0.16) {
                \hbox{\cite{10.1145/3474085.3475575}\phantom{.}}
            };
            \node[inner sep=5pt,rectangle,fill=none,draw=none] at (0.42,0.3) {
                \hbox{\cite{khandelwal2022simple}\phantom{.}}
            };
            \node[inner sep=5pt,rectangle,fill=none,draw=none] at (0.53,0.71) {
                \hbox{\cite{ramrakhya2022habitat}\phantom{.}}
            };
            \node[inner sep=5pt,rectangle,fill=none,draw=none] at (0.54,0.87) {
                \hbox{\cite{majumdar2022zson}\phantom{.}}
            };
            \node[inner sep=5pt,rectangle,fill=none,draw=none] at (0.60,0.22) {
                \hbox{\cite{yu2023l3mvn}\phantom{.}}
            };
            \node[inner sep=5pt,rectangle,fill=none,draw=none] at (0.69,0.22) {
                \hbox{\cite{Du2023ObjectGoalVN}\phantom{.}}
            };
            \node[inner sep=5pt,rectangle,fill=none,draw=none] at (0.78,0.22) {
                \hbox{\cite{yin2024sgnav}\phantom{.}}
            };
            \node[inner sep=5pt,rectangle,fill=none,draw=none] at (0.83,0.81) {
                \hbox{\cite{yokoyama2024hm3d}\phantom{.}}
            };
            \node[inner sep=5pt,rectangle,fill=none,draw=none] at (0.91,0.88) {
                \hbox{\cite{yu2024trajectory}\phantom{.}}
            };
            
            \node[inner sep=5pt,rectangle,fill=none,draw=none] at (0.45,0.92) {
                \hbox{\cite{kim2023topological}\phantom{.}}
            };
            \node[inner sep=5pt,rectangle,fill=none,draw=none] at (0.65,0.12) {
                \hbox{\cite{sun2023fgprompt}\phantom{.}}
            };
            \node[inner sep=5pt,rectangle,fill=none,draw=none] at (0.76,0.88) {
                \hbox{\cite{li2024memonav}\phantom{.}}
            };
            \node[inner sep=5pt,rectangle,fill=none,draw=none] at (0.80,0.12) {
                \hbox{\cite{sakaguchi2024object}\phantom{.}}
            };
            
            \node[inner sep=5pt,rectangle,fill=none,draw=none] at (0.19,0.81) {
                \hbox{\cite{gan2020look}\phantom{.}}
            };
            \node[inner sep=5pt,rectangle,fill=none,draw=none] at (0.29,0.22) {
                \hbox{\cite{chen2021semantic}\phantom{.}}
            };
            \node[inner sep=5pt,rectangle,fill=none,draw=none] at (0.47,0.16) {
                \hbox{\cite{paul2022avlen}\phantom{.}}
            };
            \node[inner sep=5pt,rectangle,fill=none,draw=none] at (0.63,0.94) {
                \hbox{\cite{younes2023catch}\phantom{.}}
            };
            \node[inner sep=5pt,rectangle,fill=none,draw=none] at (0.90,0.11) {
                \hbox{\cite{yang2024rila}\phantom{.}}
            };
        \end{scope}
    \end{tikzpicture}
    \caption{Timeline of the historical development of navigation tasks and their representative approaches. Different colors represent different navigation tasks, showing how the field has evolved from simpler point-goal navigation to more complex, multimodal navigation paradigms.}
    \label{fig:timeline}
\end{figure*}
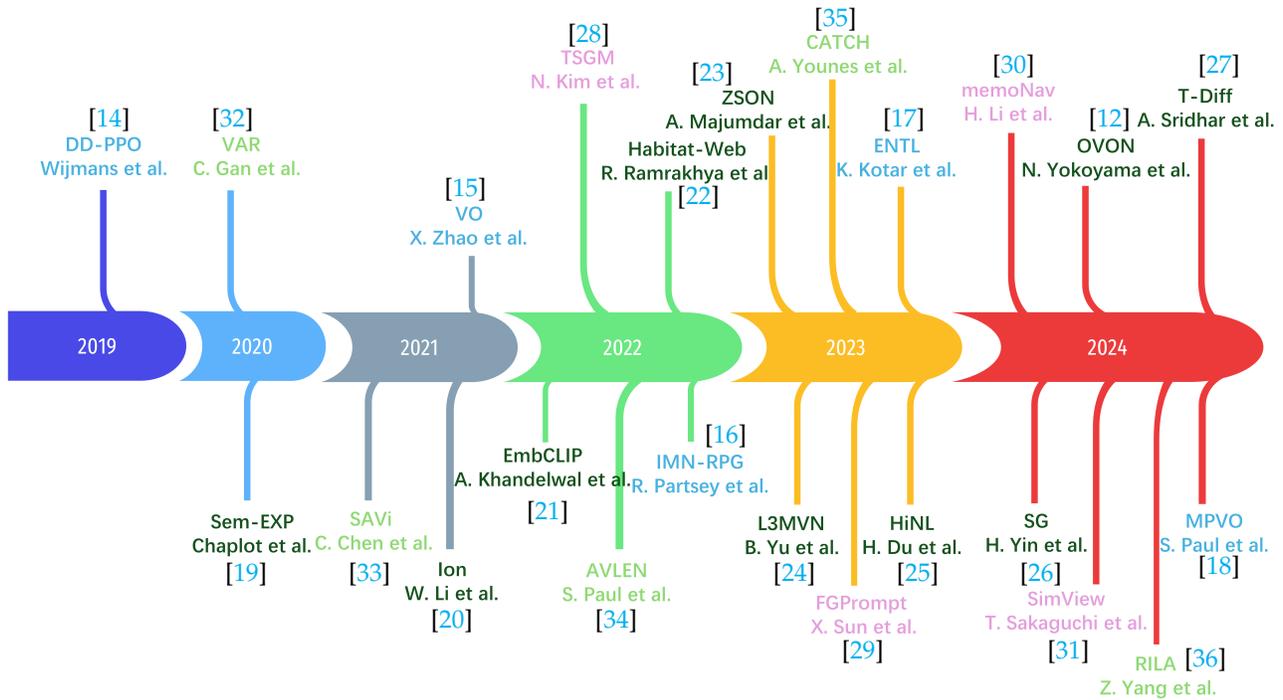
\subsection{Categorization with Inference Domains}
We organize this survey around the concept of inference domains that serve as the main environmental reasoning during navigation \cite{chaplot2020learning,wijmans2019dd,kim2023topological}. Inference domains represent the fundamental approaches to the way agents process and use information, providing a unified lens through which to analyze methods in diverse navigation tasks \cite{anderson2018evaluation,batra2020objectnav,chen2020soundspaces}. This organization offers several advantages: (i) it reveals cross-task insights as similar inference domains appear across different navigation tasks; (ii) it exposes the historical progression from explicit to implicit representations; (iii) it facilitates direct comparison of approaches sharing computational foundations and it helps identify promising research avenues applicable to multiple navigation paradigms \cite{yang2024rila,sridhar2024nomad}.

Our survey identifies and analyzes six primary inference domains that span various navigation tasks. Latent map-based methods construct and maintain explicit environmental representations, typically as 2D occupancy grids or 3D semantic maps, integrating geometric and semantic information through differentiable projections \cite{hartley2003multiple,gupta2017cognitive} to support path planning while maintaining spatial consistency \cite{sethian1996fast,ramakrishnan2022poni}. Implicit representation approaches leverage end-to-end learning to develop navigational capabilities without explicit mapping \cite{wijmans2019dd,du2021vtnet}, typically employing recurrent neural networks \cite{cho2014learning,schmidhuber1997long} or transformers \cite{vaswani2017attention,dosovitskiy2020image} to maintain an internal state that captures environmental information. Graph-based methods represent environments as relational graphs with nodes corresponding to locations, objects, or regions \cite{johnson2015image,yang2018graph}, enabling sophisticated reasoning about spatial relationships and scene layout patterns through graph neural networks \cite{kipf2016semi,scarselli2008graph,pal2021learning}. Linguistic domain approaches harness large language models (LLMs) \cite{brown2020language,achiam2023gpt} to incorporate common sense knowledge and semantic reasoning into navigation decisions through prompt-based decision-making. Embedding-based methods utilize pretrained visual-language models like CLIP \cite{radford2021learning} to establish semantic connections between visual observations and language descriptions, enabling zero-shot generalization to novel targets. Finally, the emerging diffusion model-based domain leverages denoising diffusion probabilistic models \cite{ho2020denoising,song2020score,sohl2015deep} to address navigation challenges through generative modeling approaches that include synthesizing coherent trajectory sequences, generating plausible semantic maps of unexplored areas and developing unified diffusion policies for exploration-exploitation\cite{saharia2022photorealistic,poole2022dreamfusion}.

\subsection{Contribution and Organization}
This survey makes several key contributions to the embodied AI navigation literature:
\begin{itemize}
    \item Provides the first comprehensive analysis of goal-oriented navigation approaches organized by inference domains across multiple task paradigms.
    \item Identifies common computational patterns that transcend specific navigation tasks, offering insights into fundamental principles of embodied reasoning.
    \item Systematically compares the strengths and limitations of different inference domains for various navigation scenarios.
    \item Highlight emerging trends in multimodal integration, particularly the convergence of visual, linguistic, and audio processing for enhanced navigation capabilities.
\end{itemize}

The remainder of this survey is organized as follows. Section~\ref{sec:preliminary} establishes preliminary concepts, formal definitions, datasets, and evaluation metrics common to navigation tasks. Sections~\ref{sec:pointnav} through~\ref{sec:audionav} provide in-depth analyses of Point Goal Navigation, Object Goal Navigation, Image Goal Navigation, and Audio Goal Navigation, respectively, with each section categorizing approaches by their primary inference domains. Section~\ref{sec:discussion} discusses emerging trends, challenges, and future research directions, while Section~\ref{sec:conclusion} concludes the survey with key information.

\section{Preliminary}
\label{sec:preliminary}
In this section, we provide a concise introduction to problem formulation, key topics, aiming to enhance comprehensive understanding of our work.

\subsection{Historical Development of Navigation Tasks}
The field of embodied AI navigation has evolved significantly over time, with various navigation tasks emerging to address different challenges and capabilities. Figure \ref{fig:timeline} presents a chronological overview of the development of major navigation task representative approaches.

This historical progression illustrates how navigation tasks have become increasingly complex, incorporating more modalities and challenging scenarios over time. The evolution reflects broader trends in embodied AI research, including the shift toward more naturalistic interactions, the integration of multiple sensory inputs, and the development of agents capable of understanding and following high-level semantic instructions.

\begin{table*}[ht]
\centering
\renewcommand{\arraystretch}{1}

\begin{tabular}{p{3cm}|m{2.3cm}<{\centering}|m{2.3cm}<{\centering}|m{2.3cm}<{\centering}|m{2.3cm}<{\centering}}
\toprule
\textbf{Task} & \textbf{PointNav} & \textbf{ImageNav} & \textbf{ObjectNav} & \textbf{Audio-GoalNav} \\ \midrule

\textbf{Description} & Navigate to a specific 3D point in space. & Navigate to a location matching a visual image. & Navigate to a specific object. & Navigate to sound sources. \\ \hline

\textbf{Sensory Inputs} & Visual (RGB, Depth, LiDAR, Odometry) & Visual (RGB/RGB-D) & Visual (Object recognition) & Visual (RGB-D) and binaural audio \\ \hline

\textbf{Task Goal} & Reach a specific location (coordinates). & Reach a location that matches a given image. & Find and reach a particular object. & Locate and navigate toward sound sources. \\ \hline

\textbf{Primary Challenges} & Path planning, obstacle avoidance & Visual matching, viewpoint invariance & Semantic understanding, search strategy & Sound localization, acoustic ambiguity, reverberation \\ \hline

\textbf{Success Criteria} & Agent within threshold distance of target point & Agent view matches goal image & Agent within threshold distance of target object & Agent reaches sound source location \\ \bottomrule
\end{tabular}
\caption{Comparison of Key Goal-Oriented Navigation Tasks.}
\label{tab:compared_tasks}
\end{table*}
\subsection{Formal Definition of Navigation Tasks}
Navigation tasks in embodied AI can be formalized as a decision-making process where an agent must reach a specified goal in an unknown environment through a sequence of actions. We present a mathematical framework applicable to all navigation modalities.

Let $\mathcal{E}$ represent the set of all possible environments, where each $E \in \mathcal{E}$ defines a physical space with its geometry, objects, and potential sound sources. The agent operates with the following:

\begin{itemize}
    \item \textbf{State space} $\mathcal{S}$: The set of all possible agent states, where each state $s_t \in \mathcal{S}$ includes the agent's position, orientation, and other relevant internal variables at time $t$.  
    \item \textbf{Observation space} $\mathcal{O}$: The set of all possible sensory observations. For various tasks, this may include RGB images $o_t^{rgb}$, depth maps $o_t^{depth}$, audio signals $o_t^{audio}$, or relative coordinates $o_t^{coord}$, with the full observation $o_t \in \mathcal{O}$ being the combination of available modalities. 
    \item \textbf{Action space} $\mathcal{A}$: The set of all possible actions, typically including $\mathcal{A}{=}\{\text{move\_forward}, \text{turn\_left}, \text{turn\_right}, \cdots, \text{DONE}\}$, where ``DONE'' signals task completion.
    \item \textbf{Goal space} $\mathcal{G}$: The set of possible navigation targets, which differs by task type:
    \begin{itemize}
        \item For PointNav: $g \in \mathbb{R}^2$ representing relative coordinates $(\Delta x, \Delta y)$;
        \item For ObjectNav: $g \in \mathcal{C}$ where $\mathcal{C}$ is the set of object categories;
        \item For ImageNav: $g \in \mathcal{I}$ where $\mathcal{I}$ is the set of goal images;
        \item For AudioGoalNav: $g \in \mathcal{A}ud$ where $\mathcal{A}ud$ is the set of audio signals or sources.
    \end{itemize}
\end{itemize}

An instance of a navigation task $\tau = (E, g, s_0)$ consists of an environment $E \in \mathcal{E}$, a goal $g \in \mathcal{G}$, and an initial state $s_0 \in \mathcal{S}$. The agent's objective is to learn a policy $\pi: \mathcal{O} \times \mathcal{G} \rightarrow \mathcal{A}$ that maps the current observation and the goal to an action, maximizing the expected success rate while minimizing the path length.

\subsection{Navigation Datasets}
Recent advances in embodied AI navigation research have been supported by increasingly sophisticated simulation environments and datasets. These environments vary significantly in scale, visual fidelity, and complexity, each offering different advantages for training and evaluating navigation agents. In the following, we provide a systematic comparison of the main navigation datasets and simulators.

\subsubsection{Dataset Scale and Coverage}
The Habitat-Matterport 3D (HM3D) dataset~\cite{ramakrishnan2021habitat} represents the largest collection with 1,000 building-scale reconstructions covering 112.5k $m^2$ of navigable area and 365.42k $m^2$ of total floor area. This significantly surpasses previous datasets such as Gibson~\cite{xia2018gibson} (571 scenes, 81.84k $m^2$ navigable area) and Matterport3D~\cite{chang2017matterport3d} (90 scenes, 30.22k $m^2$ navigable area). The Gibson dataset includes a high-quality subset (``Gibson 4+'') containing 106 scenes with fewer reconstruction artifacts but covering only 7.18k $m^2$ of navigable area.

AI2-THOR~\cite{kolve2017ai2} provides several dataset options including iTHOR(120 manually designed rooms), RoboTHOR~\cite{deitke2020robothor} (89 apartment-style scenes) and ProcTHOR-10k~\cite{deitke2022procthor} (10,000 procedurally generated houses). These environments are synthetic rather than scan-based that allows for infinite procedural generation of new environments.Smaller-scale datasets include ScanNet~\cite{dai2017scannet} (1,613 scenes but primarily room-scale regions totaling 10.52k $m^2$ navigable area) and Replica~\cite{straub2019replica} (18 high-fidelity scenes with just 0.56k $m^2$ navigable area).

\subsubsection{Navigation Complexity and Scene Clutter}
Navigation complexity (measured as the ratio between geodesic and Euclidean distances) varies significantly between datasets~\cite{ramakrishnan2021habitat}. Matterport3D offers the highest complexity (17.09), followed by Gibson (14.25) and HM3D (13.31). Room-scale datasets such as RoboTHOR (2.06) and ScanNet (3.78) present lower navigational challenges.

Scene clutter (the ratio between the mesh area near navigable regions and navigable space) is highest in RoboTHOR (8.2) and HM3D (3.90), with Gibson (3.14) and MP3D (2.99) offering less cluttered environments.

\subsubsection{Visual Fidelity and Reconstruction Quality}
HM3D provides superior visual fidelity with the lowest FID score (20.5) when compared to real images~\cite{ramakrishnan2021habitat}. Gibson 4+ (27.4), full Gibson (39.3), and MP3D (43.8) follow in visual quality. Synthetic environments like RoboTHOR show significantly higher FID scores (157.6).

Reconstruction artifacts such as missing surfaces, holes, or untextured regions are least common in HM3D, with 560 scenes having fewer than 5\% of views exhibiting artifacts, compared to only 175 such scenes in Gibson~\cite{xia2018gibson}. ScanNet shows the highest percentage of reconstruction defects, while Replica offers high quality but in a limited number of environments.

\subsubsection{Audio-Visual Capabilities: SoundSpaces}
SoundSpaces~\cite{chen2020soundspaces} introduces audio-visual navigation capabilities to embodied AI through realistic acoustic simulations. The platform provides audio renderings for Matterport3D (90 buildings) and Replica (18 scenes), complementing their visual data with spatial audio information. Furthermore, it evolved to SoundSpaces 2.0~\cite{chen2022soundspaces}, which introduces fast rendering, continuous spatial sampling, configurable acoustic parameters, and cross-environment generalizability. The platform simulates key acoustic phenomena, including direct sound paths, early reflections, reverberation effects, and spatial audio rendering using head-related transfer functions (HRTF), enabling agents to navigate using visual and auditory information.

\begin{table}[ht]
    \renewcommand{\arraystretch}{1.2}
    \small  
    \centering
    \caption{Comparison of Major Navigation Simulators.}
    \begin{tabular}{p{1.6cm}|p{1.8cm}|p{1.8cm}|p{1.8cm}}
    \toprule
    \textbf{Feature} & \textbf{Habitat} & \textbf{AI2-THOR} & \textbf{SoundSpaces} \\
    \midrule
    Image Quality & Photorealistic (3D scans) & Near photo-realistic (synthetic) & Compatible with Habitat \\
    \hline
    Compatible Datasets & MP3D, Gibson, HM3D & iTHOR, RoboTHOR, ProcTHOR & MP3D, Replica \\
    \hline
    Inputs & RGB-D, GPS+\newline Compass & RGB-D & RGB-D+\newline spatial audio \\
    \hline
    Action Space & Move, turn, look, stop & Move, turn, look, stop & Navigation + audio \\
    \bottomrule
    \end{tabular}
    \label{tab:simulatorComparison}
\end{table}
\subsubsection{Simulator Comparison}
The three primary simulation frameworks used with these datasets are Habitat~\cite{savva2019habitat}, AI2-THOR~\cite{kolve2017ai2}, and SoundSpaces as described in Table \ref{tab:simulatorComparison}.

AI2-THOR offers photorealistic environments based on 3D scans of real-world spaces, while Habitat offers higher-speed rendering and larger-scale photorealistic environments. SoundSpaces extends Habitat with audio capabilities, making it uniquely positioned for multisensory navigation research.

\subsubsection{Performance on Navigation Tasks}
Research shows that navigation performance scales almost linearly with the size of the dataset~\cite{ramakrishnan2021habitat}. HM3D-trained agents achieve superior cross-dataset generalization when tested across Gibson, MP3D, or HM3D environments.

With SoundSpaces, audio significantly enhances navigation performance~\cite{chen2020soundspaces, chen2022soundspaces}, allowing agents to:
\begin{itemize}
    \item Detect spatial geometry through audio reflections;
    \item Better track sounding targets;
    \item Navigate effectively when visual information is limited;
    \item Develop integrated audio-visual spatial representations.
\end{itemize}

Both Gibson and MP3D datasets can be enhanced with semantic annotations (from 3DSceneGraph\cite{armeni20193d} for Gibson, and directly from MP3D), allowing conversion to semantic maps data as Poni\cite{ramakrishnan2022poni} does. These semantically rich environments are particularly valuable for ObjectNav tasks that require object recognition and scene understanding.

\subsection{Evaluation Metrics}
Navigation performance is assessed using standardized metrics that evaluate success, efficiency, and precision:

\subsubsection{Success Metrics}
\begin{itemize}
    \item \textbf{Success Rate (SR)}: Percentage of episodes where the agent successfully reaches the goal within a specified distance threshold and declares the completion of the task.
    \item \textbf{Success weighted by Path Length (SPL)}~\cite{anderson2018evaluation}: Combines success with path efficiency, computed as described in the Formal Definition subsection.
    \begin{equation}
    \text{SPL} = \frac{1}{N}\sum_{i=1}^{N}S_i\frac{L_i}{\max(P_i,L_i)},
    \end{equation}   
    where $S_i$ indicates success (1 if successful, 0 otherwise), $L_i$ is the shortest path length and $P_i$ is the agent's actual path length for episode $i$.
    \item \textbf{Soft SPL (SoftSPL)}: Relaxes the binary success criterion with a continuous measure of goal proximity, useful for evaluating near-successes.
\end{itemize}

\subsubsection{Distance and Efficiency Metrics}
\begin{itemize}
    \item \textbf{Distance to Goal (DTG)}: Average final distance between the agent and the goal position.
    \item \textbf{Distance to Success (DTS)}: Minimum additional distance required for the agent to reach the success threshold.
    \item \textbf{Navigation Error (NE)}: Average distance between the agent's final position and the goal location.  
\end{itemize}
\begin{table*}[htbp]
    \renewcommand{\arraystretch}{1}
    \centering
    \caption{Models for Point Goal Navigation Categorized by Inference Domain (*: useful for ObjectNav)}
    \begin{tabular}{c|c|c|c|c|m{2.0cm}<{\centering}|m{2.3cm}<{\centering}|m{1.8cm}<{\centering}}
    \toprule
    \textbf{Inference Domain} & \textbf{Year} & \textbf{\#} & \textbf{Method} & \textbf{Venue} & \textbf{Architecture Type} & \textbf{Datasets} & \textbf{Metric} \\
    \midrule
    \multirow{5}{*}{\textbf{\begin{tabular}[c]{@{}c@{}}Latent Map\end{tabular}}} 
    & 2020 & \cite{chaplot2020learning} & ANM & ICLR & Modular & Gibson, MP3D & SR, SPL \\
    & 2022 & \cite{li2022comparison} & LSP-UNet & CoRL & Modular & MP3D & SR, SPL, SoftSPL \\
    & 2022 & \cite{georgakis2022uncertainty} & UPEN & ICRA & Modular & MP3D, Gibson & SR, SPL \\
    & 2022 & \cite{lee2022moda} & MoDA & ECCV & Modular & Gibson, MP3D & SR, SPL \\
    \midrule
    \multirow{15}{*}{\textbf{\begin{tabular}[c]{@{}c@{}}Implicit\\Representation\end{tabular}}} 
    & 2019 & \cite{gordon2019splitnet} & SplitNet & ICCV & Modular & MP3D, Gibson & SR, SPL \\
    & 2019 & \cite{wijmans2019dd} & DD-PPO & ICLR & Modular & Gibson, MP3D & SR, SPL \\
    & 2020 & \cite{weihs2020allenact} & DD-PPO Depth & Arxiv & Modular & iTHOR, RoboTHOR, Gibson & SR, SPL \\
    & 2021 & \cite{zhao2021surprising} & VO & ICCV & Modular & Gibson & SR, SPL \\
    & 2021 & \cite{ye2021auxiliary} & Attention Fusion & CoRL & Modular & Gibson & SR,SPL \\
    & 2021 & \cite{desai2021auxiliary} & Efficient Learning & WACV & End-To-End & Gibson & SPL, SP \\
    & 2021 & \cite{datta2021integrating} & Ego-Localization &PMLR & Modular & Gibson & SR, SPL \\
    & 2021 & \cite{chattopadhyay2021robustnav} & RobustNav* & ICCV & End-To-End & RoboTHOR, Gibson & SR, SPL \\
    & 2022 & \cite{dwivedi2022navigation} & iSEE* & CVPR & End-To-End & iTHOR & SR, SPL \\
    & 2022 & \cite{partsey2022mapping} & IMN-RPG & CVPR & Modular & Gibson, MP3D, HM3D & SR, SPL, SoftSPL, GD \\
    & 2023 & \cite{kotar2023entl} & ENTL* & ICCV & End-To-End & RoboTHOR & SR, SPL \\
    & 2024 & \cite{paul2024mpvo} & MPVO & ECCV & Modular & Gibson & SR, SPL, SoftSPL, RPE, ATE \\
    \bottomrule
    
    \multicolumn{8}{l}{\footnotesize $^1$ SR:Success Rate; SPL:Success weighted by Path Length; GD:Goal Distance; RPE:Mean Relative Pose Error;}\\
    \multicolumn{8}{l}{\footnotesize ATE:Mean Absolute Trajectory Error; SP:Success Path}\\
    
    \multicolumn{8}{l}{\footnotesize $^2$ MP3D: Matterport3D Dataset; HM3D: Habitat-Matterport 3D Dataset}
    \end{tabular}
    \label{tab:TablePointNav}
\end{table*}
\subsubsection{Multi-goal Navigation Metrics}
\begin{itemize}
    \item \textbf{Progress (PR)}: The fraction of goals successfully achieved. For single-goal tasks, this equals the Success Rate (SR).
    \item \textbf{Progress weighted by Path Length (PPL)}~\cite{anderson2018vision}: Evaluates navigation efficiency for multi-goal scenarios, computed as $\text{PPL} = \frac{1}{E}\sum_{i=1}^{E}\text{Progress}_i\frac{l_i}{\max(p_i,l_i)}$, where $E$ is the total number of test episodes, $l_i$ is the shortest path distance to the final goal via intermediate goals, and $p_i$ is the agent's actual path length.
\end{itemize}

\subsubsection{Specialized Audio Navigation Metrics}
\begin{itemize}
    \item \textbf{Sound Navigation Efficiency (SNE)}~\cite{younes2023catch}: Evaluates how efficiently an agent navigates toward dynamic sound sources.
    \item \textbf{Dynamic SPL (DSPL)}~\cite{younes2023catch}: Adaptation of SPL for moving targets, accounting for target position changes.
    \item \textbf{Success weighted by Number of Actions (SNA)}~\cite{chen2020soundspaces}: Success rate weighted by inverse number of actions.      
    \item \textbf{Success rate when silent (SWS)}~\cite{chen2022soundspaces}: Success rate when the agent is not provided with audio cues.
\end{itemize}
\section{Point Goal Navigation}
\label{sec:pointnav}
PointNav tasks involve directing an agent to a target location by given relative coordinates$(\triangle x,\triangle y)$ and without any prior knowledge of the layout~\cite{anderson2018evaluation}. The task is framed as a path-planning problem where the agent must compute a sequence of actions to reach the target from a random starting position. The primary challenge in PointNav is using egocentric sensory inputs(mostly visual data like RGB-D,GPS/Compass) to determine the location of the agent to estimate distance and plan a path whithout prior map.

\subsection{Latent Map Inference Domain}
Latent map inference domain construct explicit spatial representations that enable agents to reason about their environment through learned mapping techniques. These methods primarily focus on constructing occupancy or semantic representations from egocentric observations, allowing agents to perform path planning and exploration without prior environment knowledge. Unlike implicit methods that indirectly encode spatial information, these approaches explicitly model the geometric properties of the environment.

Early mapping approaches like ANM \cite{chaplot2020learning} introduced modular frameworks combining Neural SLAM for map construction, global policies for frontier-based exploration, and local policies for short-term action execution. This architecture was further refined in LSP-Unet \cite{li2022comparison}, which employed U-Net architectures \cite{ronneberger2015u} to estimate frontier properties from partial semantic maps, enabling more sophisticated path planning through Bellman equations. Addressing uncertainty in spatial reasoning, UPEN \cite{georgakis2022uncertainty} leveraged ensemble learning with multiple UNet models to generate variance-based uncertainty maps, guiding exploration toward informative regions while maintaining robustness through Bayesian map updating \cite{lakshminarayanan2017simple}. For domain adaptation challenges, MoDA \cite{lee2022moda} applied style transfer networks (CycleGAN \cite{zhu2017unpaired}) to translate between noisy and clean maps, enabling pre-trained navigation agents to operate effectively in visually corrupted environments.

These mapping-based approaches provide agents with explicit spatial understanding necessary to navigate efficiently without prior environment knowledge. By constructing geometric representations from sensor data, they enable agents to reason about free space, obstacles, and optimal paths to goals, forming a fundamental component of modern PointNav systems that rely on explicit spatial modeling.

\subsection{Implicit Representation Learning Inference Domain}
Implicit representation learning inference domain encode spatial understanding without constructing explicit maps, instead relying on neural network parameters to implicitly capture environmental structure. These methods learn latent representations that enable navigation through feature encoding in network weights rather than through geometric modeling. Three primary strategies define this domain: visual odometry (VO), auxiliary task learning, and policy-based approaches.

Visual odometry methods estimate agent movement through sensory observations to maintain location awareness. VO \cite{zhao2021surprising} employs a ResNet-18 backbone to estimate SE(2) transformations between consecutive RGB-D frames, using action-specific submodels and geometric invariance losses to handle sensor noise. MPVO \cite{paul2024mpvo} enhances this approach by incorporating motion priors, using a training-free geometric module to estimate coarse pose that guides a learned model for refinement. Notably, IMN-RPG \cite{partsey2022mapping} demonstrates that explicit mapping is unnecessary by combining self-supervised VO with reinforcement learning, using egomotion predictions to maintain agent-centric pose estimates that guide an LSTM-based policy.

Auxiliary task approaches enhance representation learning through complementary objectives. SplitNet \cite{gordon2019splitnet} decouples perception from policy by training a visual encoder on geometric tasks (depth mapping, surface normal estimation, egomotion calculation) and a gradient-isolated LSTM policy network, enabling both sim-to-sim and task-to-task transfer. Attention Fusion \cite{ye2021auxiliary} combines predictive coding, inverse dynamics, and temporal distance estimation through a shared encoder with parallel GRU modules and attention-based feature fusion. Similarly, Efficient Learning \cite{desai2021auxiliary} accelerates policy convergence through depth prediction, inverse dynamics modeling, and path length regression, connecting these auxiliary tasks to a central GRU policy network.

Policy-based methods focus on end-to-end training of recurrent neural architectures. DD-PPO \cite{wijmans2019dd} redefined scalability in embodied RL through decentralized distributed training, addressing convergence issues with high-dimensional inputs. This framework was extended in DD-PPO Depth \cite{weihs2020allenact} with depth-specific enhancements and curriculum learning, achieving comparable success with 10x less training data. Ego-Localization \cite{datta2021integrating} and RobustNav \cite{chattopadhyay2021robustnav} incorporate LSTM/GRU architectures to process sequential observations, while iSEE \cite{dwivedi2022navigation} uses GRU policies that integrate GPS and visual inputs to encode spatial progress. Recent transformer-based approaches like ENTL \cite{kotar2023entl} unify world modeling, localization, and imitation learning through spatio-temporal sequence prediction with VQ-GAN token prediction.

These implicit approaches demonstrate that successful navigation can be achieved without explicit mapping, often with greater computational efficiency and generalization capabilities as spatial understanding emerges through task-oriented learning and representation encoding within network parameters.
\section{Object Goal Navigation}
\label{sec:objectnav}
Object Navigation (ObjectNav) tasks~\cite{anderson2018evaluation,batra2020objectnav} require an agent to locate and navigate to a specific object within an unseen environment, such as "find a cupboard." Unlike PointNav, which is based purely on spatial coordinates, ObjectNav is driven by semantic understanding, where the agent must infer the object's position based on semantic cues. With advancements in deep learning and reinforcement learning (RL), ObjectNav is commonly tackled using approaches that can be summarized into three categories: \begin{enumerate} \item \textbf{Modular-based}: Caches the visual features from RGB-D observations into a semantic map and learns a policy to predict actions. \item \textbf{End-to-End}: Directly maps the observation features to align the relationship between the map features and the semantic meaning of the object. \item \textbf{Zero-shot}: Utilizes multimodal large language models (MLLMs) or language models (LLMs) to infer semantic relevance. \end{enumerate}
\begin{table*}[htbp]
    \renewcommand{\arraystretch}{1}
    \centering
    \caption{Models for Object Navigation Categorized by Inference Domain}
    \begin{tabular}{c|c|c|c|c|m{2.0cm}<{\centering}|m{2.3cm}<{\centering}|m{1.8cm}<{\centering}}
    \toprule
    \textbf{Inference Domain} & \textbf{Year} & \textbf{\#} & \textbf{Method} & \textbf{Venue} & \textbf{Architecture Type} & \textbf{Datasets} & \textbf{Metric} \\
    \midrule
    \multirow{11}{*}{\textbf{\begin{tabular}[c]{@{}c@{}}Latent Map\end{tabular}}}
    & 2020 & \cite{chaplot2020object} & Sem-EXP & NeurIPS & Modular & MP3D & SR, SPL \\
    & 2020 & \cite{HISNav} & HISNav & IEEE Access & Modular & HISNav & SPL \\
    & 2022 & \cite{georgakis2021learning} & L2M & ICLR & Modular & MP3D & SR, SPL, SSPL, DTS \\
    & 2022 & \cite{zhu2022navigating} & Distance Map & IROS & Modular & MP3D & SR, SPL \\
    & 2022 & \cite{ramakrishnan2022poni} & PONI & CVPR & Modular & MP3D & SR, SPL \\
    & 2023 & \cite{Zhai_2023_ICCV} & PEANUT & ICCV & Modular & MP3D, HM3D & SR, SPL \\
    & 2023 & \cite{kwon2023renderable} & RNR-Map & ICLR & Modular & MP3D & SR, SPL, SSPL, DTS \\
    & 2023 & \cite{zhang20233d} & 3D-Aware & CVPR & Modular & Gibson, MP3D & SR, SPL, SSPL, DTS \\
    & 2023 & \cite{10341959} & LocCon & IROS & Modular & HM3D & SR, SPL, DTS \\
    & 2023 & \cite{staroverov2023skill} & SkillFusion & MDPI & Modular & HM3D & SR, SPL, SSPL \\
    & 2023 & \cite{chen2023not} & StructNav & RSS & Modular & Gibson & SR, SPL, SSPL \\
    & 2023 & \cite{chang2023goat,khanna2024goat} & GOAT & CVPR & Modular & HM3D-SEM & SR, SPL \\
    \midrule
    \multirow{2}{*}{\textbf{Graph}} 
    & 2021 & \cite{10.1145/3474085.3475575} & Ion & ACM & Modular & iTHOR & SR, SPL \\
    & 2021 & \cite{pal2021learning} & MJOLNIR-o & CoRL & Modular & Visual Genome (VG) & SR, SPL \\
    \midrule
    \multirow{12}{*}{\textbf{\begin{tabular}[c]{@{}c@{}}Implicit\\Representation\end{tabular}}}
    & 2021 & \cite{du2021vtnet} & VTNet & ICLR & End-To-End & AI2-Thor & SR, SPL \\
    & 2021 & \cite{zhang2021hierarchical} & HOZ & ICCV & End-To-End & RoboTHOR & SR, SPL, SAE \\
    & 2021 & \cite{kulhanek2021visual} & DRL & IEEE Access & End-To-End & Real-World & SR, DTG \\
    & 2022 & \cite{ramrakhya2022habitat} & Habitat-web & CVPR & End-To-End & MP3D & SR, SPL \\
    & 2022 & \cite{fukushima2022object} & OMT & ICRA & End-To-End & iTHOR & SR, SPL \\
    & 2023 & \cite{song2023learning} & DITA & ICML & End-To-End & iTHOR & SR, SPL \\
    & 2023 & \cite{yadav2023offline} & Offline & ICLR & End-To-End & Gibson, MP3D, HM3D & SR, SPL \\
    & 2023 & \cite{yadav2023ovrl} & OVRL & Arxiv & End-To-End & HM3D & SR, SPL \\
    & 2023 & \cite{Du2023ObjectGoalVN} & HiNL & CVPR & End-To-End & i-THOR, RoboTHOR & SR, SPL \\
    & 2023 & \cite{10341827} & RIM & IROS & End-To-End & MP3D, HM3D & SR, SPL, SSPL \\
    & 2023 & \cite{ramrakhya2023pirlnav} & PIRLNav & CVPR & End-To-End & MP3D, HM3D-Sem & SR, SPL \\
    & 2024 & \cite{yokoyama2024hm3d} & OVON & IROS & End-To-End & MP3D, HM3D, procTHOR & SR, SPL \\
    \midrule
    \multirow{6}{*}{\textbf{Linguistic}}
    & 2023 & \cite{yu2023l3mvn} & L3mvn & IROS & Modular & Gibson, HM3D & SR, SPL, DTG \\
    & 2023 & \cite{wang2024find} & DDN & NeurIPS & Modular & ProcThor & SR, SPL, SSR \\
    & 2023 & \cite{zhou2023esc} & ESC & ICML & Zero-shot & MP3D, HM3D, RoboThor & SR, SPL, SSPL \\
    & 2024 & \cite{zhang2024imagine} & SGM & CVPR & Modular & Gibson, MP3D & SR, SPL, DTS \\
    & 2024 & \cite{kuang2024openfmnav} & OpenFMNav & NAACL & Zero-shot & HM3D & SR, SPL \\
    & 2024 & \cite{wu2024voronav} & Voronav & ICML & Zero-shot & HM3D, HSSD & SR, SPL \\
    & 2024 & \cite{yin2024sgnav} & SG & NeurIPS & Zero-shot & Gibson, MP3D, HM3D, RoboTHOR & SR, SPL \\
    & 2024 & \cite{cai2024bridging} & PixNav & ICRA & Zero-shot & HM3D & SR, SPL, DTS \\
    \midrule
    \multirow{5}{*}{\textbf{CLIP Embeddings}}
    & 2022 & \cite{khandelwal2022simple} & EmbCLIP & CVPR & Zero-shot & HM3D & SPL, SR, SSPL, GD \\
    & 2022 & \cite{al2022zero} & ZSEL & CVPR & Zero-shot & MP3D, HM3D & SR, SPL \\
    & 2023 & \cite{gadre2023cows} & COWs & CVPR & Zero-shot & MP3D, RoboThor & SR, SPL \\
    & 2023 & \cite{majumdar2022zson} & ZSON & NeurIPS & Zero-shot & MP3D, HM3D & SR, SPL \\
    & 2024 & \cite{xu2024aligning} & AKGVP & ICRA & Zero-shot & iTHOR & SR, SPL, DTS \\
    \midrule
    \textbf{BLIP Embeddings} & 2023 & \cite{yokoyama2024vlfm} & Vlfm & ICRA & Zero-shot & Gibson, MP3D, HM3D & SR, SPL \\
    \midrule
    \multirow{2}{*}{\textbf{Diffusion Learning}}
    & 2024 & \cite{ji2024diffusion} & DAR & CVPR & Modular & Gibson, MP3D, HM3D & SR, SPL, DTS \\
    & 2024 & \cite{yu2024trajectory} & T-Diff & NeurIPS & Modular & MP3D, Gibson & SR, SPL, DTS \\
    \bottomrule
    
    \multicolumn{8}{l}{\footnotesize $^1$ SR:Success Rate; SPL:Success weighted by Path Length; GD(Goal Distance): The average distance between the agent and the goal}\\
    \multicolumn{8}{l}{at the end of an episode; SAE: Success weighted by Action Efficiency; DTG:Distance to Goal; DTS: Distance to Success}\\
    
    \multicolumn{8}{l}{\footnotesize $^2$ MP3D: Matterport3D Dataset; HM3D: Habitat-Matterport 3D Dataset; HM3DSem: Habitat-Matterport 3D Semantics Dataset;}\\
    \multicolumn{8}{l}{\footnotesize Gibson: Gibson Environment for Perception and Navigation Dataset; iTHOR: Interactive THOR Environment;}\\
    \multicolumn{8}{l}{\footnotesize RoboTHOR: Robot THOR Environment; ProcTHOR: Procedurally Generated THOR Environment; AI2-Thor: The House Of inteRactions;}\\
    \multicolumn{8}{l}{\footnotesize HSSD: Home Storage Scene Dataset; HISNav: HIStory-aware Navigation Dataset; VG: Visual Genome Dataset;}
    \end{tabular}
    \label{tab:TableObjectNav}
\end{table*}
\subsection{Categorization of Object Navigation Models}
Object Navigation (ObjectNav) is a fundamental task in embodied agent, where an agent must navigate to a specified object within an environment. Recent advancements in ObjectNav have been categorized into three primary methodological approaches: Modular-Based, End-To-End, and Zero-Shot methods as shown in Tab.\ref{tab:TableObjectNav}. In the following subsection, we provide detailed decription of method categories.
\begin{figure}[t]
    \centering
    \includegraphics[width=\linewidth]{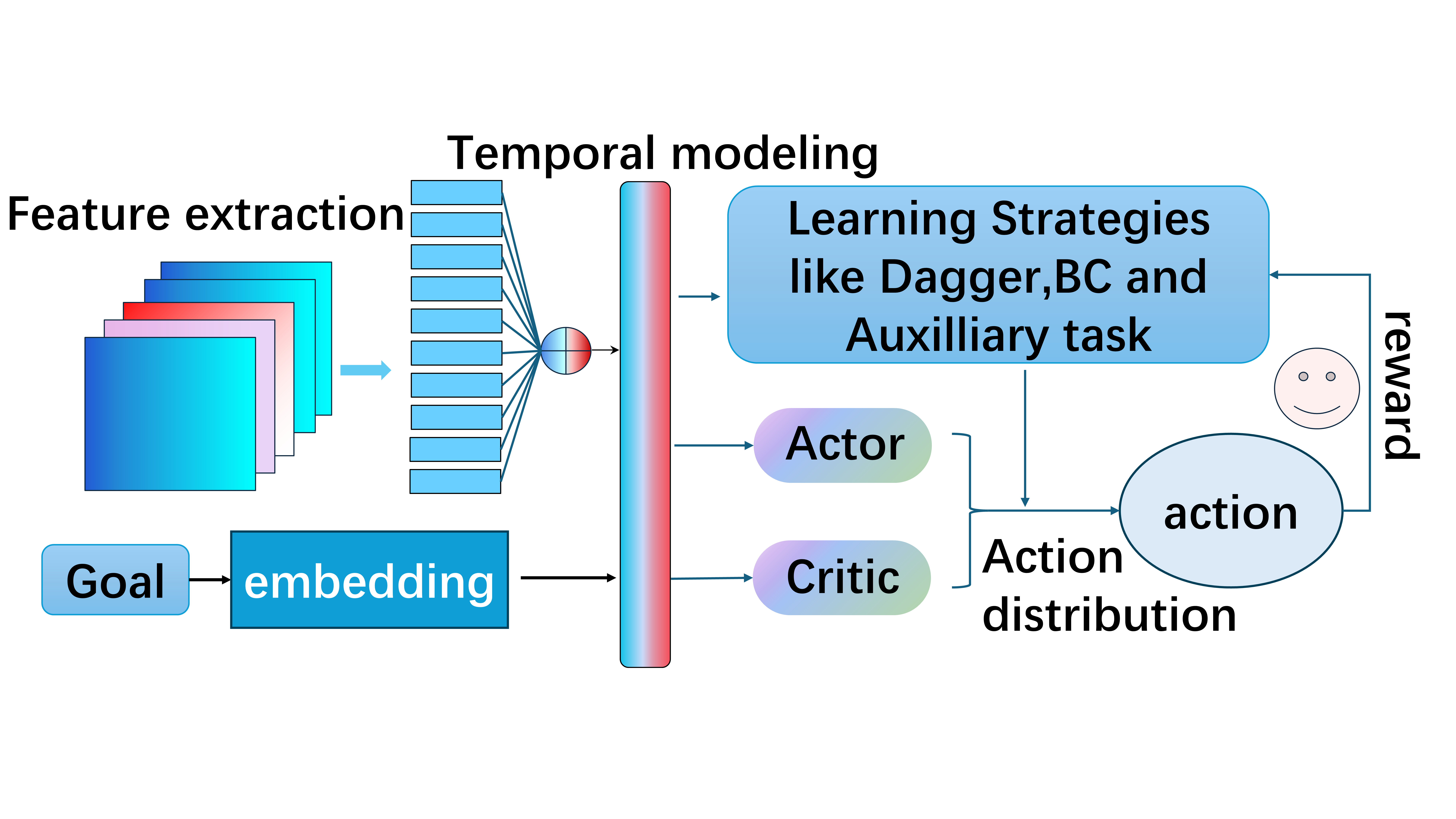} 
    \caption{Implicit Representation Learning Inference Domain: Behavioral cloning learns directly from expert trajectories; DAgger iteratively collects expert feedback; and auxiliary tasks create additional feedback signals to improve reward utilization.}
    \label{fig:e2e_approaches}
\end{figure}
\subsection{End-To-End Methods}

End-To-End methods directly learn navigation policies from raw sensory inputs to actions without explicit intermediate representations~\cite{zhu2017target,mirowski2018learning}. Unlike modular approaches that separate mapping and planning, these methods learn navigation capabilities through unified architectures that implicitly encode environmental understanding~\cite{gupta2017cognitive,chaplot2020learning}. The visual representation component processes raw sensory data through neural networks~\cite{he2016deep,lecun2015deep}, while the policy learning component translates these representations into actions through reinforcement learning or imitation learning paradigms~\cite{mnih2016asynchronous,schulman2017proximal,ross2011reduction}. Recent advances have focused on enhancing both components through transformer architectures~\cite{vaswani2017attention,dosovitskiy2020image}, memory-augmented neural networks~\cite{graves2016hybrid,wayne2018unsupervised} and computationally efficient learning strategies~\cite{yadav2023ovrl}. This end-to-end approach offers advantages in terms of architectural simplicity and adaptation capability although challenges remain in terms of sample efficiency and generalization to novel environments~\cite{levine2016end,savva2019habitat}.

\subsubsection{Implicit Representation Learning Inference Domain}
Implicit representation learning inference domain represents a comprehensive approach to navigation that jointly learns environmental representations and action policies through end-to-end training. This domain is characterized by several key aspects that emerge consistently across implementations: (1) the integration of deep neural architectures (CNNs, RNNs, Transformers) to process visual observations and encode both spatial and temporal information~\cite{lecun2015deep,vaswani2017attention}; (2) the combination of multiple learning strategies including reinforcement learning algorithms~\cite{mnih2016asynchronous,schulman2017proximal}, imitation learning approaches~\cite{ross2011reduction} and self-supervised representation learning~\cite{he2022masked}; (3) the incorporation of memory modules and attention mechanisms that maintain historical context and enable reasoning about past states~\cite{graves2016hybrid,wayne2018unsupervised}; and (4) the dynamic updating of latent representations that implicitly encode both semantic and spatial relationships while guiding policy decisions. These approaches serve as unified frameworks that bridge perception, representation and action through end-to-end learning paradigms. Their effectiveness stems from the ability to simultaneously learn rich environmental representations and robust navigation policies without requiring explicit environmental mapping.

Recent object navigation models leverage implicit representations to jointly learn environmental understanding and action policies. VTNet \cite{du2021vtnet} utilizes spatial-aware descriptors with DETR \cite{carion2020end} for object detection and a pre-training scheme to associate visual features with navigation signals. Similarly, DRL approaches \cite{kulhanek2021visual,kulhanek2019vision} employ neural architectures combining convolutional layers with LSTM \cite{schmidhuber1997long} for sequential processing, trained via algorithms like PAAC \cite{clemente2017efficient}.

DITA \cite{song2023learning} introduces a Judge Model for termination decisions alongside DRL, using MJOLNIR-o \cite{pal2021learning} as backbone and leveraging GNN to process object relations from Visual Genome \cite{krishna2017visual}. Memory-augmented approaches include RIM \cite{10341827}, which constructs spatial grids of latent vectors updated recursively through transformers \cite{waswani2017attention}, and OMT \cite{fukushima2022object}, which integrates Object-Scene Memory with transformer-based inference.

Human demonstrations have proven valuable for navigation learning. Habitat-web \cite{ramrakhya2022habitat} collects demonstrations via web-based teleoperation, showing that imitation learning outperforms reinforcement learning in ObjectNav tasks. Building on this, PIRLNav \cite{ramrakhya2023pirlnav} employs a two-stage approach combining behavior cloning pretraining with RL finetuning, using ResNet50 \cite{he2016deep} and GRU layers.

State-modeling approaches include HiNL \cite{Du2023ObjectGoalVN}, which explicitly models relationships among historical navigation states through History-aware State Estimation and History-based State Regularization. OVRL \cite{yadav2023ovrl} integrates Vision Transformers \cite{alexey2020image} with compression layers and LSTM networks, demonstrating positive scaling laws through MAE pretraining \cite{he2022masked}.

Transformer-based methods like OVON \cite{yokoyama2024hm3d} utilize frozen SigLIP \cite{zhai2023sigmoid} encoders for visual and textual inputs. Their findings demonstrate that DAgger \cite{ross2011reduction} pretraining with frontier exploration trajectories followed by RL finetuning achieves superior performance, while integration with object detectors like OWLv2 \cite{minderer2024scaling} improves generalization to unseen categories.

\subsection{Modular-Based Methods}

\begin{figure}[t]
    \centering
    \includegraphics[width=\linewidth]{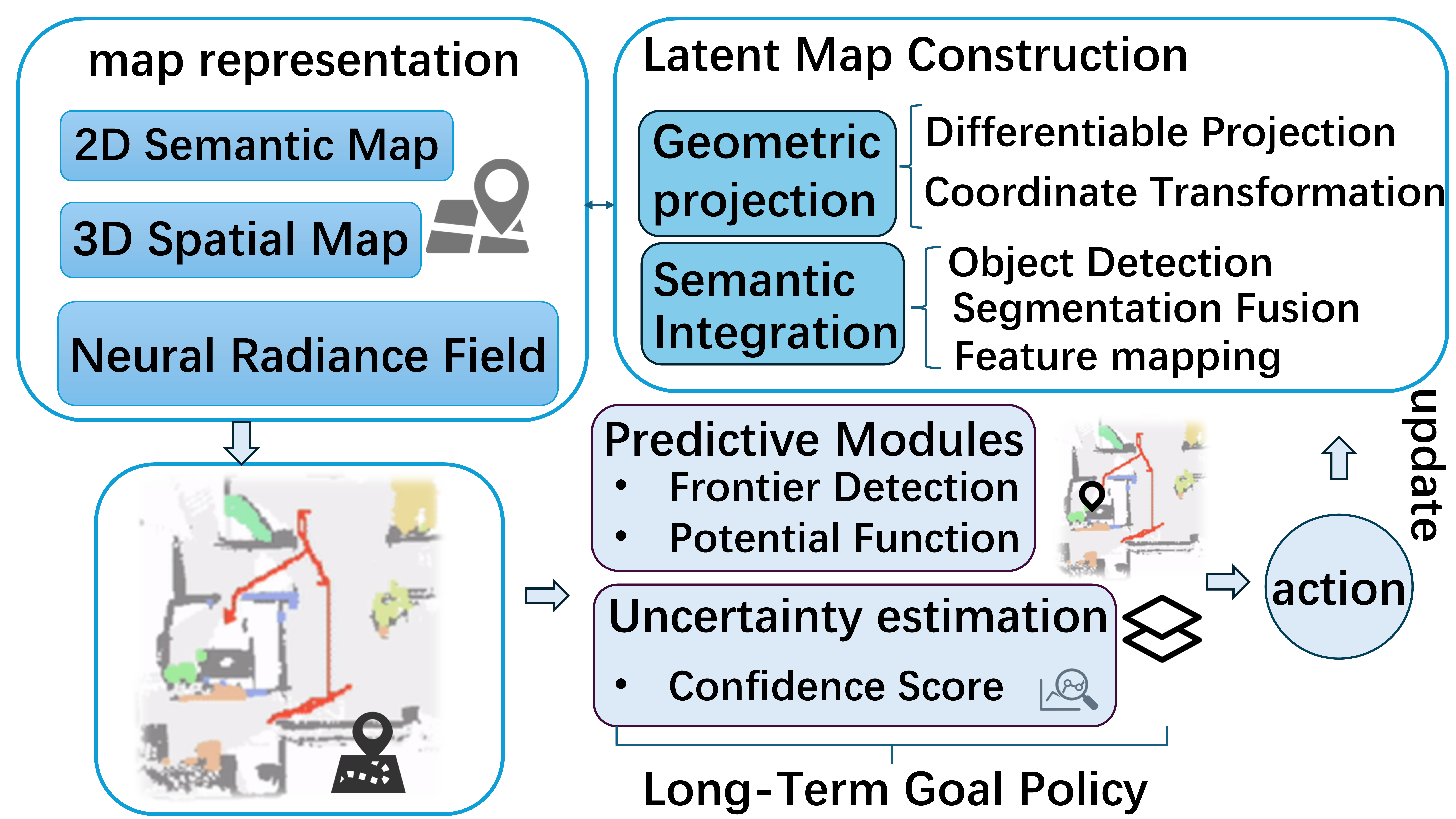}
    \caption{Latent Map Based Inference Domain: This domain constructs environmental representations that combine geometric and semantic information through mapping modules, using these maps as active inference domains to guide navigation decisions through path planning and policy modules.}
    \label{fig:modular_approaches}
\end{figure}

Modular methods decompose the ObjectNav task into distinct modules: mapping, policy, and path planning. The mapping module constructs a representation of the environment, typically as a latent map or a graph, which is then used by the policy module to generate a long-term goal. The path planning module subsequently guides the agent to this goal. latent map representations can be further divided into those that use only observed data and those that incorporate predictions, such as completing partial maps or predicting target locations. On the other hand, Graph-based methods represent the environment using topological maps or semantic graphs, leveraging relational information to inform navigation decisions. While modular methods offer advantages in terms of interpretability and potential for real-world transfer, they also present challenges due to the diversity of mapping modules and the lack of a unified framework, making it difficult to compare different approaches effectively.

\subsubsection{Latent Map Based Inference Domain}
The Latent Map Based Inference Domain represents a sophisticated approach to environmental understanding and navigation, where agents construct and maintain internal representations that combine geometric, semantic and spatial information. These maps serve not merely as passive memory structures but as active inference domains that guide decision-making through various techniques such as potential functions\cite{ramakrishnan2022poni}, uncertainty estimation, or distance metrics\cite{sethian1996fast}. 

Semantic map construction forms the foundation for several ObjectNav approaches. Sem-EXP \cite{chaplot2020object} constructs an episodic semantic map using differentiable projections \cite{chaplot2020learning} and employs Mask R-CNN \cite{he2017mask} for object detection with a goal-oriented semantic policy for long-term navigation planning. Similarly, PEANUT \cite{Zhai_2023_ICCV} employs PSPNet \cite{zhao2017pyramid} to generate semantic segmentation masks projected onto a top-down map, with its prediction model forecasting target probabilities in unexplored areas. L2M \cite{georgakis2021learning} takes this further by actively learning to predict semantic maps beyond the agent's field of view, employing uncertainty estimates from segmentation model ensembles \cite{lakshminarayanan2017simple} to guide long-term goal selection within a POMDP framework.

Potential functions and distance metrics are leveraged by several approaches to optimize exploration and goal-directed behavior. PONI \cite{ramakrishnan2022poni} introduces a dual potential function network where one potential guides frontier-based exploration \cite{yamauchi1997frontier} while the other enables goal-directed navigation through geodesic distance metrics. The Distance map approach \cite{zhu2022navigating} combines a bird's-eye view semantic map with a predicted distance map that estimates geodesic distances to potential targets, leveraging learned spatial correlations between objects \cite{wang2018efficient} to prioritize exploration of areas likely to minimize distance to target objects.

Advanced 3D representations enhance environmental understanding for navigation. 3D-Aware \cite{zhang20233d} constructs detailed 3D maps through online semantic point fusion \cite{dai2017scannet,zhang2020fusion}, using a corner-guided exploration policy alongside a category-aware identification policy optimized via PPO \cite{schulman2017proximal}. RNR-Map \cite{kwon2023renderable} employs latent codes from image observations embedded into grid cells, transformable into neural radiance fields \cite{mildenhall2021nerf} for image rendering, with inverse projection \cite{hartley2003multiple} mapping visual information for location matching.

Self-supervised and training-free approaches reduce dependency on labeled data. LocCon \cite{10341959} introduces a two-stage process using contrastive learning \cite{chen2020simple} to fine-tune semantic segmentation \cite{chen2017deeplab} on multi-viewpoint images, followed by navigation policy development using potential functions over map frontiers \cite{yamauchi1997frontier}. StructNav \cite{chen2023not} presents a training-free approach using Visual SLAM \cite{mur2015orb} to construct a structured scene representation, enhancing frontier-based exploration with semantic scoring from language models (BERT \cite{devlin2018bert}, CLIP \cite{radford2021learning}). HISNav \cite{HISNav} addresses real-world deployment challenges through hierarchical reinforcement learning with ORB-SLAM2 \cite{mur2017orb} for localization and a custom dataset based on Habitat \cite{savva2019habitat} for pre-training before real-world deployment.

\begin{figure}[t]
    \centering
    \includegraphics[width=\linewidth]{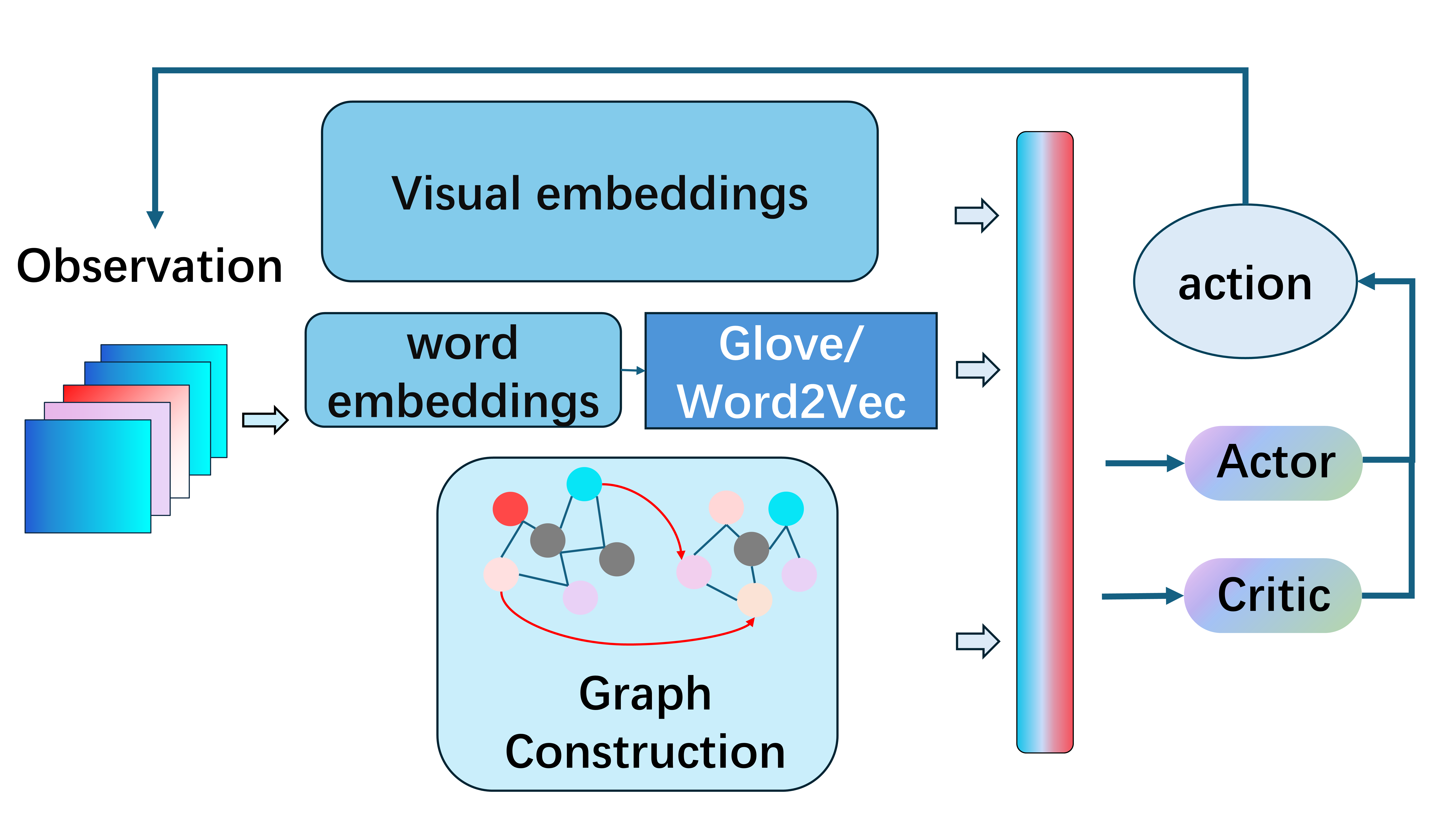}
    \caption{Graph Based Inference Domain: This domain constructs hierarchical graph representations that capture relationships between environmental elements at different levels of abstraction, leveraging these structured representations to enable semantic reasoning and more efficient navigation decisions through graph-based algorithms.}
    \label{fig:graph_approaches}
\end{figure}

\subsubsection{Graph Based Inference Domain}
The Graph Based Inference Domain represents a sophisticated approach to navigation that leverages structured graph representations to capture complex relationships between environmental elements and guide decision-making processes. This domain is characterized by the hierarchical organization of spatial and semantic information, typically representing scenes, zones, and objects as interconnected nodes through Graph Convolutional Neural Networks\cite{kipf2016semi}; the integration of visual features with word embeddings to create rich node representations; and the use of contextual information to enhance navigation through policy optimization. These graph-based representations enable agents to reason about spatial relationships and scene layout patterns, facilitating more efficient navigation strategies. The effectiveness of this approach lies in its ability to decompose complex navigation tasks into hierarchical sub-problems\cite{zhang2021hierarchical} while maintaining semantic consistency through the graph structure and allowing for robust decision-making in partially observed environments.

Graph-based approaches leverage structured representations to capture complex environmental relationships for effective navigation. Ion \cite{10.1145/3474085.3475575} employs an Instance-Relation Graph with GCNs to process Faster-RCNN \cite{ren2016faster} features, while formulating navigation as a POMDP \cite{kaelbling1998planning} optimized through A3C \cite{mnih2016asynchronous}. HOZ \cite{zhang2021hierarchical} introduces a hierarchical structure with scene, zone, and object nodes to capture spatial layouts \cite{zuo2016learning}, planning optimal paths between zones using an LSTM network. MJOLNIR-o \cite{pal2021learning} constructs a knowledge graph using a Contextualized Graph Network \cite{kipf2016semi} that embeds object relationships based on spatial and semantic proximities from Visual Genome \cite{krishna2017visual}, with reward shaping that encourages exploration around parent objects while integrating observation and contextual information.

    \begin{figure}[t]
        \centering
        \includegraphics[width=\linewidth]{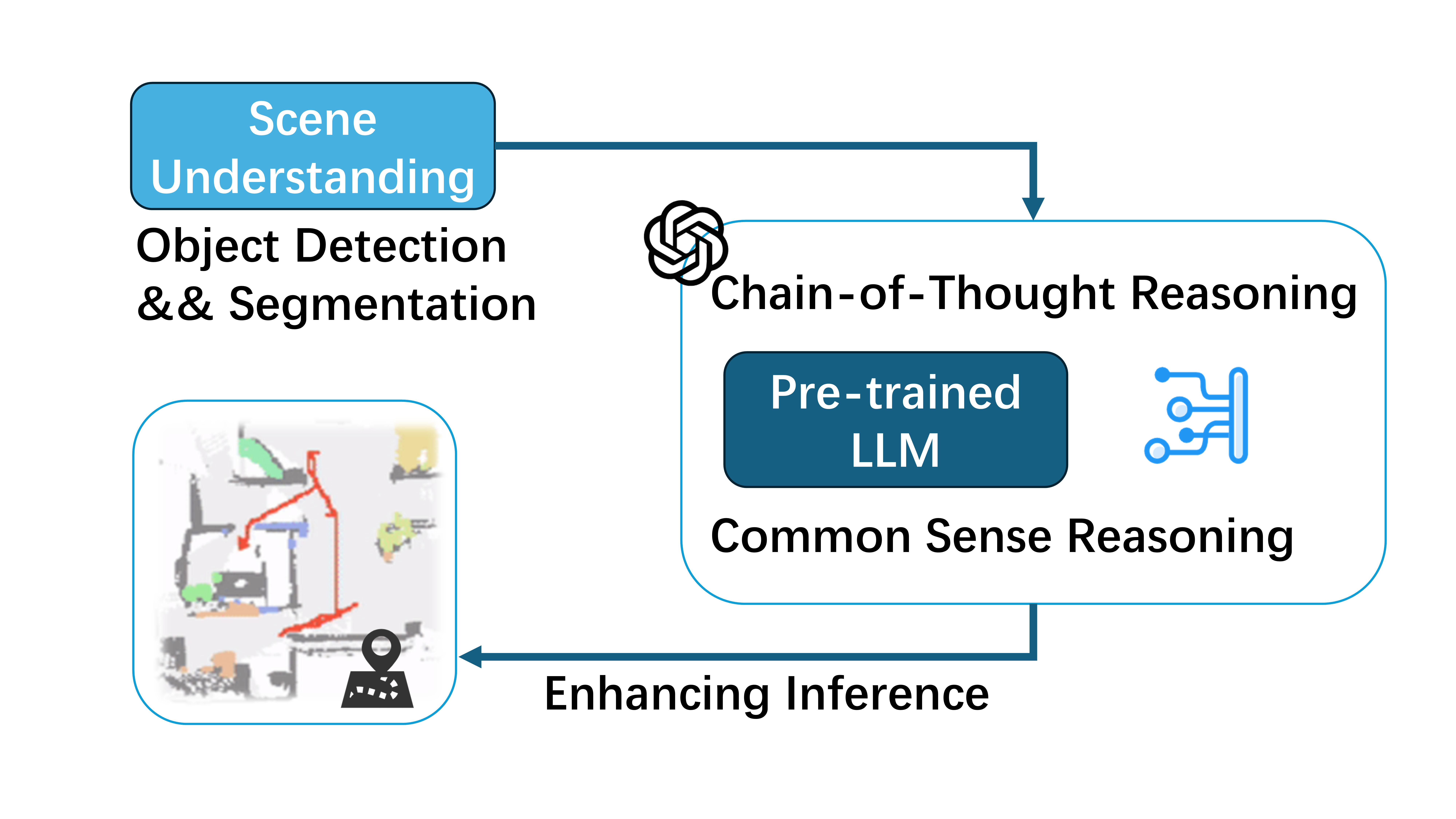} 
        \caption{Linguistic Inference Domain: This domain leverages large language models to enhance navigation through semantic reasoning, providing common-sense knowledge about object relationships and spatial layouts while enabling sophisticated decision-making through natural language understanding.}
        \label{fig:linguistic_approaches}
    \end{figure}
\subsubsection{Linguistic Inference Domain}
The Linguistic Inference Domain represents an advanced approach to navigation that harnesses the semantic understanding capabilities of large language models (LLMs)\cite{zhu2023minigpt,achiam2023gpt,devlin2018bert,chowdhery2023palm,thoppilan2022lamda} to enhance environmental reasoning and decision-making processes. This domain is characterized by several key aspects that emerge consistently across implementations: the integration of common-sense knowledge extracted from LLMs to understand object-environment relationships and spatial layouts, the use of language models for generating and evaluating descriptive representations of environmental states and navigation goals and the application of linguistic reasoning to guide exploration strategies and predict unobserved regions. These language-driven approaches serve as sophisticated inference frameworks that enable agents to leverage semantic knowledge for enhanced navigation, bridging the gap between natural language understanding and spatial reasoning. The effectiveness of linguistic-based approaches lies in their ability to incorporate high-level semantic knowledge into navigation decisions while maintaining robust generalization capabilities across diverse environments and task requirements.

DDN \cite{wang2024find} employs LLM-generated descriptions to learn textual attribute features aligned with visual features through CLIP embeddings \cite{radford2021learning}, creating a textual-visual alignment mechanism for demand-conditioned navigation. L3mvn \cite{yu2023l3mvn} integrates LLMs \cite{brown2020language,devlin2018bert} with semantic map construction, implementing frontier detection \cite{yamauchi1997frontier} with two selection paradigms: a zero-shot approach where LLMs evaluate frontier descriptions and a feed-forward paradigm using language model embeddings with a fine-tuned neural network. SGM \cite{zhang2024imagine} predicts unobserved environmental regions through masked modeling techniques \cite{he2022masked}, employing multi-scale training with systematically masked semantic maps and integrating visual observations with LLM knowledge (GPT-4 \cite{achiam2023gpt}, ChatGLM \cite{du2021glm}) to guide exploration through probabilistic estimates of target object locations.

\begin{figure}[t]
    \centering
    \includegraphics[width=\linewidth]{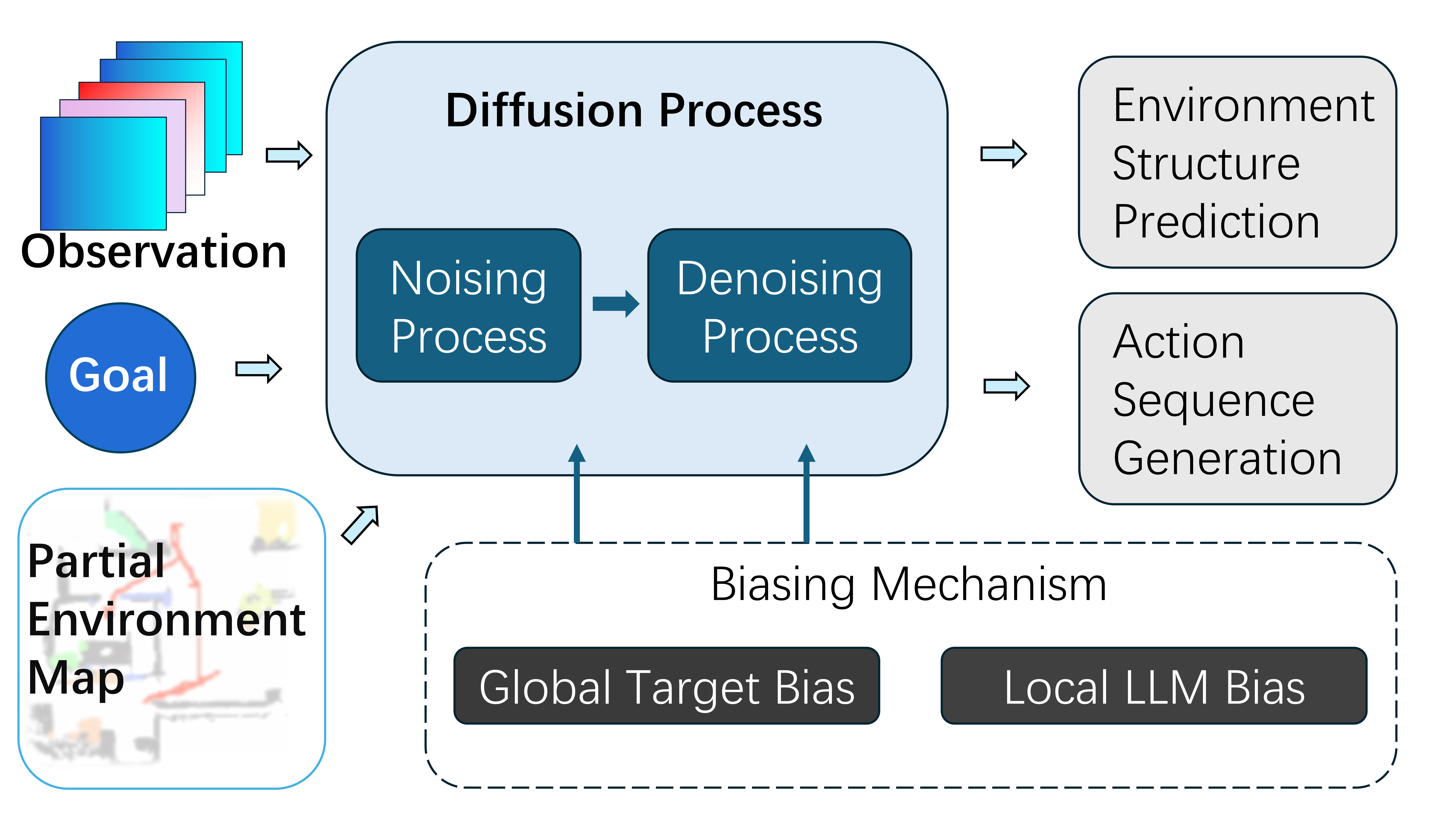} 
    \caption{Diffusion Model Based Inference Domain: This domain leverages denoising diffusion probabilistic models to simultaneously generate environmental structures and action sequences. By converting noise to structured outputs through iterative denoising, these models uniquely span both perception and action aspects of navigation.}
    \label{fig:diffusion_approaches}
\end{figure}
\subsubsection{Diffusion Model Based Inference Domain}
The Diffusion Model Based Inference Domain represents an innovative approach to navigation that leverages denoising diffusion probabilistic models\cite{ho2020denoising,song2020score,sohl2015deep} to generate and reason about both environmental structures and action sequences. This domain is characterized by several distinctive capabilities: generating plausible semantic maps of unobserved regions through conditional diffusion processes\cite{dhariwal2021diffusion,nichol2021improved}; incorporating biasing mechanisms that guide generation toward task-relevant outcomes\cite{saharia2022photorealistic}; producing multimodal action distributions that capture navigation uncertainty; and generating temporally coherent trajectory sequences~\cite{chi2023diffusion}. Unlike other inference domains that focus primarily on either perception or action, diffusion models uniquely span both aspects of navigation, enabling agents to simultaneously reason about environmental structure and optimal trajectories. These approaches effectively bridge the gap between partial observations and complete scene understanding, while maintaining semantic consistency with observed features.

DAR \cite{ji2024diffusion} generates plausible semantic maps of unexplored areas through two key components: a "global target bias" guiding generation toward likely target object locations and a "local LLM bias" incorporating common-sense knowledge about object relationships \cite{brown2020language}, enabling semantic reasoning while maintaining consistency with observed environmental structure \cite{karras2022elucidating}. T-Diff \cite{yu2024trajectory} reconceptualizes embodied navigation as a generative modeling challenge by employing diffusion models to generate temporally coherent trajectory sequences conditioned on visual observations and target semantics. By decomposing navigation into trajectory generation and following phases, the system achieves superior planning capabilities through hierarchical reasoning.

\subsection{Zero-Shot Methods}

\hspace{2em}Zero-Shot methods aim to generalize navigation capabilities to unseen environments or object categories without explicit retraining, often leveraging large-scale pre-trained models and semantic embeddings.

\begin{figure}[t]
    \centering
    \includegraphics[width=\linewidth]{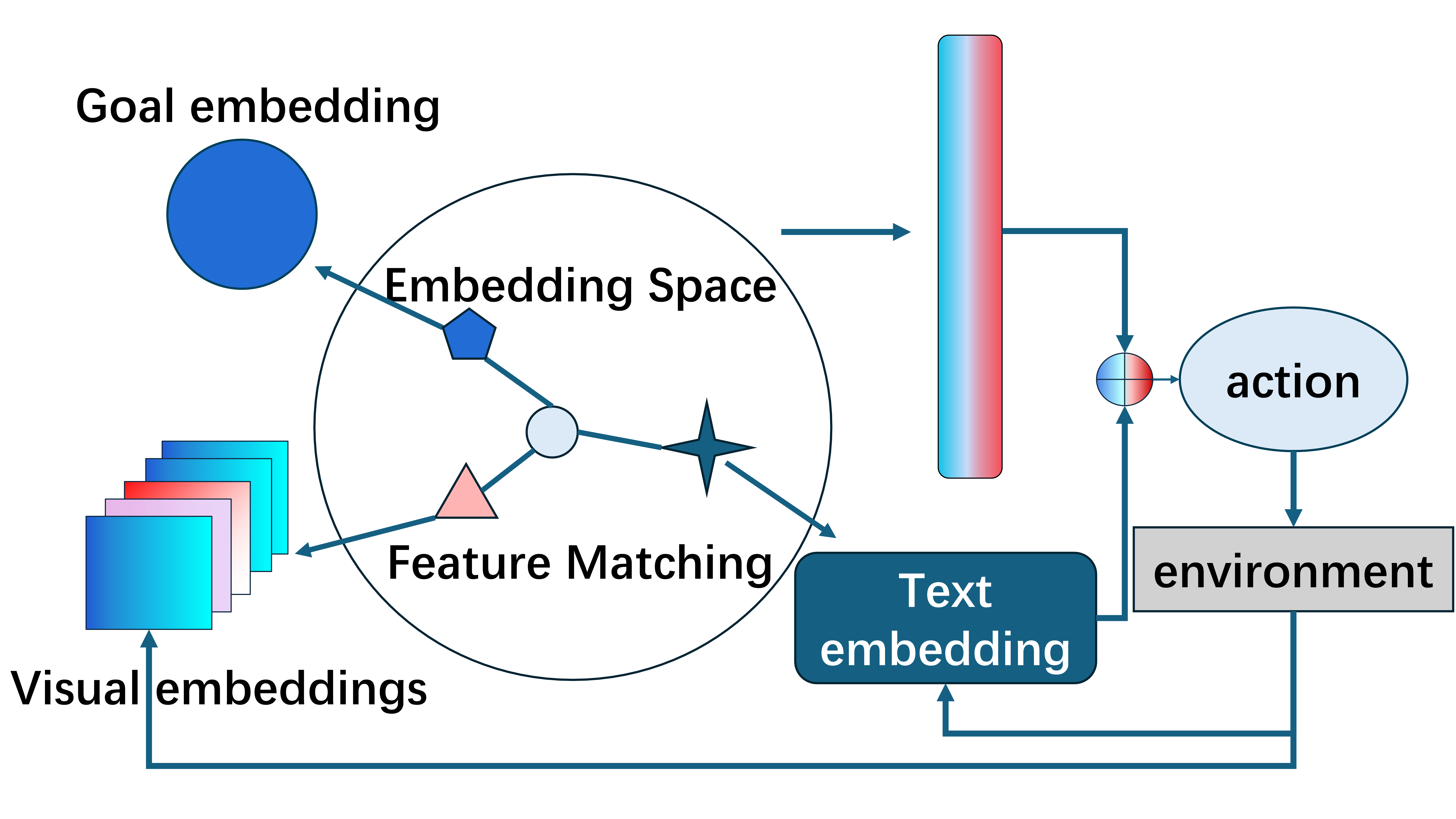}
    \caption{Embedding Based Inference Domain: This approach leverages pre-trained vision-language models to establish semantic connections between visual observations and language descriptions, enabling zero-shot generalization to unseen objects through unified embedding spaces and similarity-based reasoning.}
    \label{fig:embedding_approaches}
\end{figure}

\subsubsection{Embedding-Based Inference Domain}
The Embedding-Based Inference Domain leverages pre-trained vision-language models~\cite{radford2021learning,li2023blip} to enable zero-shot generalization across unseen objects and environments. This Domain is characterized by: frozen vision-language models establishing semantic representations~\cite{he2016deep,radford2021learning}, multi-modal embeddings enabling cross-modal understanding~\cite{chen2020simple} and frontier-based exploration enhanced by semantic similarity scores~\cite{yamauchi1997frontier}. These methods bridge visual perception and semantic understanding through unified embedding spaces~\cite{anderson2018vision,shridhar2020alfred}, Thus, it demonstrates robust generalization while maintaining computational efficiency through modular architectures.

EmbCLIP \cite{khandelwal2022simple} presents a streamlined approach using frozen CLIP ResNet-50 \cite{he2016deep,radford2021learning} embeddings processed through a GRU \cite{cho2014learning} for effective action prediction, demonstrating that these embeddings inherently capture semantic and geometric information crucial for navigation. Building on similar principles, ZSEL \cite{al2022zero} introduces a modular transfer learning framework for zero-shot generalization, combining view alignment optimization \cite{mirowski2016learning} with task augmentation techniques \cite{tobin2017domain} and establishing a joint goal embedding space that aligns various input modalities with image goals \cite{radford2021learning,chen2020simple}.

Object localization techniques enhance semantic navigation capabilities in zero-shot settings. COWs \cite{gadre2023cows} leverages CLIP embeddings \cite{radford2021learning} for language-driven navigation without prior training, integrating depth-based mapping with both frontier-based exploration \cite{yamauchi1997frontier} and learned policies. Similarly, ZSON \cite{majumdar2022zson} employs CLIP embeddings as its primary reasoning component, creating a shared semantic space for goals specified through various modalities, using ResNet-50 \cite{he2016deep} for observation processing coupled with an LSTM-based policy network to facilitate generalization to novel object categories without additional training.

Multimodal integration frameworks combine various vision-language models with exploration strategies. VLFM \cite{yokoyama2024vlfm} integrates frontier-based exploration \cite{yamauchi1997frontier} with vision-language embeddings, constructing an occupancy map from depth observations \cite{chaplot2020learning} while generating a value map using BLIP-2 \cite{li2023blip} that computes cosine similarity between observations and text prompts. The framework incorporates state-of-the-art detection models including YOLOv7 \cite{wang2023yolov7}, Grounding-DINO \cite{liu2024grounding}, and Mobile-SAM \cite{zhang2023faster} for instance segmentation, demonstrating strong zero-shot generalization capabilities while maintaining computational efficiency through its modular design.

\subsubsection{Linguistic Inference Dominance}
The Linguistic Inference Dominance leverages large language models (LLMs)~\cite{brown2020language,devlin2018bert} and vision-language models~\cite{li2022grounded,radford2021learning} for zero-shot navigation generalization. This domain integrates multi-stage reasoning where LLMs provide common-sense knowledge about spatial relationships, utilizes prompt-based architectures for structured decision-making~\cite{wei2022chain}, and combines visual perception with linguistic understanding through vision-language models for scene comprehension~\cite{anderson2018vision}. These language-driven approaches bridge perception and action through modular frameworks that separate high-level reasoning from low-level control. The effectiveness stems from decomposing navigation tasks through natural language reasoning while maintaining generalization capabilities via foundation models.

ESC \cite{zhou2023esc} bridges scene understanding and navigation decisions through three components: open-world scene understanding using GLIP \cite{li2022grounded} for prompt-based object detection, commonsense reasoning with LLMs to infer spatial relationships and guided exploration using Probabilistic Soft Logic \cite{bach2017hinge} to convert inferences into navigation strategies. OpenFMNav \cite{kuang2024openfmnav} employs multiple vision-language models in a modular framework with dedicated components for parsing instructions, scene understanding \cite{du2021glm,devlin2018bert}, object detection \cite{zhang2023faster}, and semantic mapping\cite{chaplot2020learning} with LLM-based reasoning to guide exploration strategies. 

Structural approaches facilitate navigation through environmental decomposition, with systems like Voronav \cite{wu2024voronav} using Reduced Voronoi Graphs \cite{aurenhammer1991voronoi} with LLM reasoning, SG \cite{yin2024sgnav} constructing 3D scene graphs with chain-of-thought prompting, and PixNav \cite{cai2024bridging} utilizing transformer architectures with foundation model interfaces \cite{zhang2023llama,achiam2023gpt} for high-level planning through structured prompting \cite{wei2022chain}.
\section{Image Goal Navigation}
\begin{table*}[htbp]
    \renewcommand{\arraystretch}{1}
    \centering
    \caption{Models for Image Navigation Categorized by Inference Domain}
    \begin{tabular}{c|c|c|c|c|m{2.0cm}<{\centering}|m{2.3cm}<{\centering}|c}
    \toprule
    \textbf{Inference Domain} & \textbf{Year} & \textbf{\#} & \textbf{Method} & \textbf{Venue} & \textbf{Architecture Type} & \textbf{Datasets} & \textbf{Metric} \\
    \midrule
    \multirow{6}{*}{\textbf{\begin{tabular}[c]{@{}c@{}}latent map\end{tabular}}} 
    & 2022 & \cite{wu2022image} & MIGN & IEEE & Modular & Gibson & SR, SPL \\
    & 2022 & \cite{mezghan2022memory} & MANav & IROS & End-To-End & Gibson & SR, SPL \\
    & 2023 & \cite{krantz2023navigating} & Mod-IIN & ICCV & Modular & HM3D & SR, SPL, NE \\
    & 2024 & \cite{johnson2024feudal} & FeudalNav & CVPR & Modular & Gibson & SR, SPL, DTS \\
    & 2024 & \cite{li2024memonav} & memoNav & CVPR & Modular & Gibson, MP3D & SR, SPL, PR, PPL \\
    & 2024 & \cite{jiao2024litevloc} & LiteVLoc & IROS & Modular & Gibson, MP3D & SR, SPL \\
    \midrule
    \multirow{4}{*}{\textbf{\begin{tabular}[c]{@{}c@{}}Implicit\\ Representation\end{tabular}}}
    & 2022 & \cite{bono2023end} & EmerNav & ICLR & End-To-End & Gibson, MP3D, HM3D & SR, SPL \\
    & 2023 & \cite{wasserman2023last} & SLING & CoRL & Modular & Gibson, MP3D & SR, SPL \\
    & 2023 & \cite{sun2023fgprompt} & FGPrompt & NeurIPS & End-To-End & Gibson, MP3D, HM3D & SR, SPL \\
    & 2024 & \cite{wang2024enhancing} & NUENav & IROS & End-To-End & Gibson & SR, SPL, DTS \\
    & 2024 & \cite{sakaguchi2024object} & SimView & IROS & Modular & HM3D & SR, SPL \\
    \midrule
    \textbf{Graph} & 2022 & \cite{kim2023topological} & TSGM & CoRL & Modular & Gibson & SR, SPL \\
    \midrule
    \textbf{Diffusion Learning} & 2024 & \cite{sridhar2024nomad} & NOMAD & ICRA & Modular & GNM, SACSoN & SR, SPL \\
    \bottomrule
    
    \multicolumn{8}{l}{\footnotesize $^1$ SR:Success Rate; SPL:Success weighted by Path Length; GD:Goal Distance; NE:Navigation Error; PR:progress;}\\ 
    \multicolumn{8}{l}{PPL:progress weighted by path length}\\
    
    \multicolumn{8}{l}{\footnotesize $^2$ MP3D: Matterport3D Dataset; HM3D: Habitat-Matterport 3D Dataset}
    \end{tabular}
    \label{tab:TableImageNav}
\end{table*}
\label{sec:imagenav}
\hspace{2em}Image Goal Navigation (ImageNav)~\cite{zhu2021soon,anderson2018vision,batra2020objectnav} presents a fundamental challenge in embodied AI, requiring agents to navigate to a destination specified solely by a reference image without explicit coordinates. Unlike point-goal navigation, which relies on metric information, ImageNav demands visual reasoning capabilities to establish correspondences between current observations and goal images.

\subsection{Latent Map Based Inference Domain}
The Latent Map Based Inference Domain in ImageNav constructs and maintains explicit environmental representations to facilitate goal matching and path planning. These approaches integrate geometric and visual information to build spatial memory structures while enabling efficient retrieval of goal-relevant information.

Memory-augmented systems like MANav \cite{mezghan2022memory} enhance navigation through a sophisticated memory module with self-supervised state embedding networks and episodic memory mechanisms for previously visited states \cite{oord2018representation,pritzel2017neural}. Similarly, memoNav \cite{li2024memonav} implements three distinct memory types (short-term, long-term, and working memory) with selective forgetting mechanisms and graph attention networks \cite{vaswani2017attention} to optimize information processing.

Modular frameworks decompose the navigation challenge through specialized components. Mod-IIN \cite{krantz2023navigating} combines frontier-based exploration \cite{yamauchi1997frontier} with goal instance re-identification using SuperPoint \cite{detone2018superpoint} and SuperGlue \cite{sarlin2020superglue}, while MIGN \cite{wu2022image} integrates Neural-SLAM \cite{chaplot2020learning} with reinforcement learning for goal prediction. Hierarchical approaches like FeudalNav \cite{johnson2024feudal} create self-supervised Memory Proxy Maps \cite{pang2022unsupervised} with waypoint networks trained on human data, while LiteVLoc \cite{jiao2024litevloc} implements global and local localization modules with covisibility graphs \cite{rublee2011orb} and SLAM integration \cite{ila2009information}.

\subsection{Implicit Representation Inference Domain}
The Implicit Representation Inference Domain encodes environmental understanding without explicit maps, relying on neural network parameters to capture spatial relationships implicitly. These approaches jointly learn perception and policy through end-to-end training while maintaining computational efficiency.

EmerNav \cite{bono2023end} employs a correspondence network to estimate matching features between observations and goal images \cite{weinzaepfel2022croco}, enabling direct navigation without explicit mapping. NUENav \cite{wang2024enhancing} leverages Neural Radiance Fields (NeRF) \cite{mildenhall2021nerf} as a cognitive structure, extracting both uncertainty and spatial features to balance exploration and exploitation. 

Fine-grained approaches like FGPrompt \cite{sun2023fgprompt} enhance goal-directed reasoning through dual fusion mechanisms: Early Fusion performing pixel-level concatenation of goal and observation images, and Mid Fusion employing Feature-wise Linear Modulation \cite{perez2018film} to dynamically adjust observation encoder activations. SLING \cite{wasserman2023last} integrates neural keypoint descriptors \cite{detone2018superpoint} with perspective-n-point algorithms \cite{lepetit2009epnp} and adaptive exploration-exploitation strategies.

SimView \cite{sakaguchi2024object} employs unimodal contrastive learning between multi-view images \cite{chen2021exploring} rather than relying on pre-trained vision-language models. By recording feature vectors during exploration and using contrastive learning to increase similarity between different views of the same object, SimView outperforms multimodal approaches like CLIP \cite{radford2021learning} for instance-level identification in navigation scenarios.

\subsection{Graph Based Inference Domain}
The Graph Based Inference Domain represents environments as relational structures with nodes corresponding to distinct locations and edges capturing connectivity. These approaches enable sophisticated reasoning about spatial relationships while facilitating efficient planning through graph traversal algorithms.

TSGM \cite{kim2023topological} implements a dual-memory system comprising a topological graph for spatial representation \cite{savinov2018semi} and semantic features for visual recognition \cite{he2016deep}. The framework constructs a sparse topological map during exploration with nodes representing distinct locations \cite{chen2019behavioral}, while employing a hierarchical decision-making process for planning \cite{wortsman2019learning}. By maintaining both structural and semantic information in a graph structure, TSGM effectively balances exploration and goal-directed navigation while adapting to environment complexity.

\subsubsection{Diffusion Model Based Inference Domain}
In the context of Image Goal Navigation, diffusion models offer unique capabilities for generating navigation policies that bridge visual perception and action planning. NOMAD \cite{sridhar2024nomad} introduces a diffusion-based policy approach with a unified architecture for both exploration and goal-directed behaviors, employing goal-masking for conditional inference and a 1D conditional U-Net to capture multimodal action distributions. Unlike previous methods that use separate policies for exploration and goal-directed navigation, NOMAD's unified framework can handle both image-goal navigation and undirected exploration through its diffusion policy component, which effectively captures the inherent multimodality of navigation decisions. This approach demonstrates strong generalization capabilities by learning spatial relationships and object placement distributions while maintaining adaptability to novel environments.

\section{Audio Goal Navigation}
\label{sec:audionav}
\begin{table*}[htbp]
    \renewcommand{\arraystretch}{1}
    \centering
    \caption{Models for Audio Goal Navigation Categorized by Inference Domain}
    \begin{tabular}{c|c|c|c|c|m{2.0cm}<{\centering}|m{2.3cm}<{\centering}|c}
    \toprule
    \textbf{Inference Domain} & \textbf{Year} & \textbf{\#} & \textbf{Method} & \textbf{Venue} & \textbf{Architecture Type} & \textbf{Datasets} & \textbf{Metric} \\
    \midrule
    \multirow{3}{*}{\textbf{\begin{tabular}[c]{@{}c@{}}latent map\end{tabular}}} 
    & 2020 & \cite{gan2020look} & VAR & ICRA & Modular & Visual-Audio Room \newline (AI2-THOR + \newline Resonance Audio) & SR, SPL \\
    & 2021 & \cite{chen2020learning} & AV-WaN & ICLR & Modular & Replica\cite{straub2019replica} & SR, SPL, SNA \\
    & 2024 & \cite{chen2024sim2real} & AFP & IROS & Modular & SoundSpaces & SR, SPL, SSPL \\
    \midrule
    \multirow{7}{*}{\textbf{\begin{tabular}[c]{@{}c@{}}Implicit\\ Representation\end{tabular}}}
    & 2021 & \cite{chen2021semantic} & SAVi & CVPR & End-To-End & MP3D & SR, SPL, SNA, \newline DTG, SWS \\
    & 2022 & \cite{yu2022sound} & SAAVN & ICLR & End-To-End & SoundSpaces, \newline Replica, MP3D & SR, SPL, DTG, \newline SoftSPL, NDTG \\
    & 2022 & \cite{paul2022avlen} & AVLEN & NeurIPS & End-To-End & SoundSpaces & SR, SPL, SNA, \newline DTG, SWS \\
    & 2023 & \cite{chen2023omnidirectional} & ORAN & ICCV & End-To-End & SoundSpaces & SR, SPL \\
    & 2023 & \cite{younes2023catch} & CATCH & IEEE & End-To-End & SoundSpaces 2.0\cite{chen2022soundspaces} & SR, SPL, SNE, \newline DSPL, DSNE \\
    & 2023 & \cite{kondoh2023multi} & SDM & ICRA & End-To-End & SoundSpaces 2.0 & SR, SPL, PPL, \newline PROGRESS \\
    & 2024 & \cite{liu2024caven} & CAVEN & AAAI & End-To-End & SoundSpaces & SR, SPL, SNA, \newline DTG, SWS \\
    \midrule
    \textbf{CLIP Embedding} & 2023 & \cite{huang2023audio} & AVLMaps & ISER & Modular & SoundSpaces & SR, SPL, SWS, \newline DTG \\
    \midrule
    \textbf{Linguistic} & 2024 & \cite{yang2024rila} & RILA & CVPR & Zero-shot & SoundSpaces & SR, SPL, DTG \\
    \bottomrule
    
    \multicolumn{8}{l}{\footnotesize $^1$ SR:Success Rate; SPL:Success weighted by Path Length; DTG:Distance to Goal; SNA:Success weighted by Number of Actions;}\\
    \multicolumn{8}{l}{\footnotesize SWS:Success weighted by Time Steps; SNE:Success weighted by Navigation Efficiency; DSPL:Dynamic SPL; DSNE:Dynamic SNE;}\\
    \multicolumn{8}{l}{\footnotesize PPL:Progress Path Length;}\\
    
    \multicolumn{8}{l}{\footnotesize $^2$ MP3D: Matterport3D Dataset; HM3D: Habitat-Matterport 3D Dataset}
    \end{tabular}
    \label{tab:TableAudioGoalNav}
\end{table*}
Audio Goal Navigation requires autonomous agents to localize and navigate toward sound sources in complex environments. This task integrates spatial audio processing with visual perception and path planning, addressing challenges including reverberation, interference, signal attenuation, and ambiguous directional cues. The field has evolved from simplistic scenarios with continuous sounds to addressing intermittent audio, competing sources, and semantically meaningful acoustic events. We categorize Audio Goal Navigation approaches based on their primary inference domains.

\subsection{Latent Map Based Inference Domain}
The Latent Map Based Inference Domain in Audio Goal Navigation constructs explicit spatial-acoustic representations to guide navigation. These approaches integrate visual geometry with sound localization to build comprehensive environmental maps supporting path planning and obstacle avoidance.

VAR~\cite{gan2020look} implements a three-component architecture integrating visual perception mapping, sound localization through STFT spectrograms processed by convolutional networks, and dynamic path planning using Dijkstra's algorithm. AV-WaN \cite{chen2020learning} constructs spatial audio intensity maps \cite{gao2020visualechoes} alongside geometric representations, employing neural encoders and a GRU-based temporal integrator \cite{cho2014learning} for predicting intermediate navigation goals. The framework utilizes hierarchical reinforcement learning and structured acoustic memory \cite{henriques2018mapnet}, generating waypoints based on environmental complexity. AFP \cite{chen2024sim2real} addresses simulation-to-reality transfer through adaptive frequency processing that enhances model generalization across acoustic conditions while maintaining structured map representations.

\subsection{Implicit Representation Inference Domain}
The Implicit Representation Inference Domain encodes spatial-acoustic understanding within neural network parameters without explicit map construction. These approaches range from standard end-to-end models to sophisticated memory-augmented architectures addressing various scenarios from static to dynamic sound sources.

For static source navigation, ORAN~\cite{chen2023omnidirectional} introduces Confidence-Aware Cross-task Policy Distillation (CCPD) and Omnidirectional Information Gathering (OIG), where CCPD transfers knowledge from PointGoal navigation policies while OIG provides 360-degree awareness by integrating observations from multiple directions. 

Dynamic source navigation approaches include CATCH \cite{younes2023catch}, which extends audio-visual navigation to moving sound sources by integrating depth-constructed geometric maps \cite{gupta2017cognitive} with binaural audio inputs through specialized encoders and a GRU memory component. SDM \cite{kondoh2023multi} addresses multiple concurrent sound sources with a Sound Direction Map that dynamically localizes multiple sources utilizing temporal memory. SAAVN \cite{yu2022sound} formulates audio-visual navigation as a zero-sum two-player game between navigator and sound attacker, enabling robust policies through competitive optimization.

For semantic audio-visual navigation, SAVi \cite{chen2021semantic} enables navigation to objects based on their sporadic semantic sounds. The transformer-based architecture processes visual \cite{he2016deep} and binaural audio inputs, predicting both location and category information \cite{mousavian2019visual} while leveraging persistent memory for long-term dependencies \cite{parisotto2020stabilizing}.

Language-integrated approaches include AVLEN \cite{paul2022avlen}, which combines audio-visual navigation with natural language assistance through hierarchical reinforcement learning, and CAVEN \cite{liu2024caven}, which implements bidirectional language interaction formulated as a budget-aware partially observable semi-Markov decision process \cite{krishnamurthy2016partially}.

\subsection{Embedding Based Inference Domain}
The Embedding Based Inference Domain leverages pre-trained vision and audio models to establish semantic connections between acoustic, visual, and linguistic modalities through unified embedding spaces.

AVLMaps \cite{huang2023audio} extends audio-visual navigation by integrating natural language understanding within a comprehensive 3D spatial mapping framework. The system implements four localization modules: a visual module employing NetVLAD \cite{arandjelovic2016netvlad} and SuperPoint \cite{detone2018superpoint}, an object module utilizing open-vocabulary segmentation \cite{ghiasi2022scaling}, an area localization module, and an audio module leveraging AudioCLIP \cite{guzhov2022audioclip}. This multimodal approach creates a unified spatial representation enabling complex target specifications \cite{jatavallabhula2023conceptfusion}.

\subsection{Linguistic Inference Domain}
The Linguistic Inference Domain harnesses large language models to enhance audio-visual navigation through semantic reasoning, enabling zero-shot generalization and sophisticated decision-making without task-specific training.

RILA \cite{yang2024rila} introduces a zero-shot semantic audio-visual navigation framework leveraging large language models for environmental reasoning. The architecture comprises a perception module using pre-trained vision-language models \cite{radford2021learning,li2023blip}, an Imaginative Assistant constructing cognitive maps \cite{gupta2017cognitive}, and a Reflective Planner employing frontier-based exploration \cite{yamauchi1997frontier}. The system employs sophisticated audio localization combining distance and directional estimates weighted by signal intensity \cite{morgado2020learning}, demonstrating effective generalization across unseen environments without task-specific training.
\section{Discussion}
\label{sec:discussion}
The evolution of multimodal navigation approaches across diverse tasks reveals underlying computational patterns that transcend specific application domains. By analyzing navigation methods through the lens of inference domains—from explicit map-based representations to emerging diffusion-based generative approaches—we uncover shared mechanisms that enable agents to perceive, reason about, and traverse complex environments. These commonalities suggest fundamental principles in how embodied systems process spatial information, integrate multiple modalities, and make sequential decisions under uncertainty. Our analysis demonstrates how similar computational frameworks adapt to the unique challenges posed by different navigation paradigms, with task-specific specializations emerging from shared architectural foundations. In this section, we examine cross-task insights revealed by our inference domain framework, analyze current challenges including security vulnerabilities, and identify promising research directions that can address these limitations while advancing the field toward more capable, generalizable navigation systems.

\subsection{Cross-Task Insights: Inference Domains Across Navigation Paradigms}
Our analysis of inference domains across PointNav, ObjectNav, ImageNav, and AudioGoalNav reveals distinctive adaptation patterns that illuminate the fundamental nature of each domain. We highlight key cross-task insights below:

\subsubsection{Latent Map Based Adaptations}
Latent map approaches demonstrate a clear evolution in complexity and information content across navigation tasks. In PointNav, maps primarily encode geometric information (occupancy grids) with frontier-based exploration \cite{chaplot2020learning}, while ObjectNav maps integrate semantic object labels and likelihood distributions \cite{ramakrishnan2022poni,Zhai_2023_ICCV}. ImageNav further extends this domain by incorporating visual feature vectors within spatial structures \cite{mezghan2022memory,li2024memonav}, while AudioGoalNav maps uniquely represent sound propagation patterns alongside visual geometry \cite{gan2020look,chen2020learning}. This progression reflects how the underlying representation adapts to the increasing semantic complexity of the task while maintaining the fundamental principles of explicit spatial modeling.

\subsubsection{Implicit Representation Specialization}
Implicit representations exhibit task-specific specializations while sharing core architectural elements. PointNav approaches emphasize visual odometry and pose estimation \cite{zhao2021surprising,paul2024mpvo}, whereas ObjectNav agents incorporate object relationship modeling \cite{du2021vtnet,song2023learning}. ImageNav models focus on visual correspondence mechanisms between current and goal images \cite{sun2023fgprompt,bono2023end}, and AudioGoalNav agents develop specialized audio-visual integration strategies \cite{chen2023omnidirectional,chen2021semantic}. These adaptations highlight how similar recurrent and transformer architectures can be specialized for fundamentally different information processing requirements.

\subsubsection{Graph Based Semantic Shifts}
Graph-based approaches show significant variation in node semantics and relational structures across tasks. ObjectNav graphs typically represent object-scene relationships \cite{10.1145/3474085.3475575,zhang2021hierarchical}, with edges capturing functional and spatial connections. In contrast, ImageNav graphs like TSGM \cite{kim2023topological} represent visually distinct locations as nodes with navigability connections, creating more topological structures. This distinction reveals how the same computational framework can represent fundamentally different environmental abstractions based on task requirements.

\subsubsection{Linguistic Integration Strategies}
The linguistic inference domain demonstrates varying integration depths across navigation paradigms. While rarely used in PointNav (which requires minimal semantic understanding), ObjectNav approaches like L3mvn \cite{yu2023l3mvn} and SGM \cite{zhang2024imagine} leverage LLMs for sophisticated reasoning about object relationships and spatial layouts. In AudioGoalNav, systems like RILA \cite{yang2024rila} integrate language with audio-visual perception to reason about sound sources and their semantic properties. These approaches highlight a common pattern: language becomes more valuable as the task's semantic complexity increases.

\subsubsection{Embedding Balance and Adaptation}
Embedding-based methods show varying balances between pre-trained knowledge and task-specific adaptation. ObjectNav approaches like ZSON \cite{majumdar2022zson} and EmbCLIP \cite{khandelwal2022simple} leverage CLIP's semantic knowledge almost directly, while AudioGoalNav systems like AVLMaps \cite{huang2023audio} must carefully integrate AudioCLIP embeddings with spatial reasoning. This pattern reveals how foundation model transfer becomes more challenging as the sensory gap between pre-training and navigation increases.

\subsubsection{Diffusion Models for Environmental Synthesis}
The emerging diffusion model domain, most developed in ObjectNav with approaches like NOMAD \cite{sridhar2024nomad} and DAR \cite{ji2024diffusion}, offers particular promise for tasks requiring semantic prediction of unobserved regions. This pattern suggests that generative environmental modeling provides maximum utility for tasks with high semantic complexity and partial observability, explaining its current concentration in ObjectNav research.

\subsection{Current Challenges}
Embodied navigation has evolved dramatically over the past decade, progressing from simple geometric path planning to sophisticated multimodal systems integrating visual, linguistic, and audio understanding. Despite significant advances across multiple inference domains—from explicit map-based approaches to emerging diffusion-based generative methods—several fundamental challenges persist across navigation paradigms.
\subsubsection{Sim-to-Real Transfer}
The simulation-to-reality gap persists as a critical challenge across navigation tasks. While environments like Habitat-Matterport 3D offer impressive visual fidelity (FID scores of 20.5 compared to real images), significant discrepancies remain in physical dynamics, sensor noise characteristics, and acoustic properties. Methods like AFP have begun addressing acoustic sim-to-real gaps through frequency-adaptive techniques, but comprehensive solutions remain elusive.

Additionally, most current benchmarks evaluate discrete action spaces in relatively controlled environments, failing to capture the continuous control challenges and environmental unpredictability of real-world deployment scenarios.
\subsubsection{Multimodal Representation and Integration}
While significant progress has been made in integrating multiple modalities, optimal fusion strategies remain an open question. Current approaches often prioritize one sensory modality, with others serving in supplementary roles. For example, in AudioGoalNav, audio often provides directional cues while visual data primarily handles obstacle avoidance. Few approaches achieve truly balanced integration that leverages the complementary strengths of each modality.

The challenge extends to representation learning across modalities. Visual representations (through CNNs, ViTs) are well-developed, while audio representations for navigation remain comparatively underdeveloped. Language representation, despite advances with LLMs, still faces challenges in grounding linguistic concepts to physical environments.

\subsection{Future Work}
Based on trends observed across navigation paradigms and inference domains, we identify several promising research directions that may address current limitations while advancing the field toward more capable, generalizable navigation systems.

\subsubsection{Human-Guided Scene Generalization}
Technical approaches like domain randomization \cite{tobin2017domain} and procedural generation \cite{deitke2022procthor} offer limited solutions for novel environments. Human guidance, as shown in \cite{liu2024caven}, provides complementary support through clarification requests. Future systems should combine automated generalization with strategic human interaction. Foundation models offer generalizable visual concepts but need better spatial grounding through human demonstration data. The most promising direction involves systems that recognize their limitations and request assistance when needed, combining Linguistic mechanisms for communication, Latent Maps for spatial representation, and Diffusion Models for environmental completion from minimal guidance.

\subsubsection{Unified Multimodal Representation Learning}
While current systems often prioritize one sensory modality, with others in supplementary roles, future work should focus on truly balanced integration. Research into joint embedding spaces that preserve modality-specific strengths while enabling cross-modal reasoning will be essential. We identify particular opportunities for audio-visual-language fusion through shared tokenization approaches \cite{huang2023audio} and cross-modal attention mechanisms that dynamically weight modalities based on reliability estimates. The development of multimodal foundation models specifically trained for embodied navigation tasks represents a promising direction for establishing unified environmental understanding frameworks.
\section{Conclusion}
\label{sec:conclusion}
This survey has presented a comprehensive analysis of multimodal navigation approaches through the unifying lens of inference domains, examining how embodied agents perceive, reason about, and navigate through complex environments. As autonomous systems continue to advance toward real-world deployment, the ability to effectively integrate multiple sensory modalities—visual, audio, linguistic, and spatial that has emerged as a critical capability for robust navigation in diverse settings. Our analysis demonstrates how the field has evolved from purely geometric reasoning toward increasingly sophisticated semantic understanding, with recent approaches leveraging foundation models, diffusion-based generative techniques, and multimodal integration to achieve unprecedented performance. By categorizing navigation methods according to their primary environmental reasoning mechanisms, we have identified recurring patterns and distinctive strengths across different navigation paradigms, revealing how similar computational foundations support seemingly disparate approaches across PointNav, ObjectNav, ImageNav, and AudioGoalNav tasks. Despite significant progress, important challenges persist in multimodal representation fusion, simulation-to-reality transfer and computational efficiency. Looking forward, the convergence of multiple inference domains—combining the spatial precision of map-based approaches with the semantic richness of language models and the generative capabilities of diffusion models—holds tremendous promise for developing navigation systems capable of operating in complex, dynamic environments alongside humans. We hope this survey serves as both a comprehensive reference and a catalyst for future innovation in multimodal navigation, ultimately contributing to autonomous systems that can safely and efficiently interact with the world in ways that meaningfully augment human capabilities.
\bibliographystyle{IEEEtran}
\bibliography{ref}

\end{document}